# Rapid post-disaster infrastructure damage characterisation enabled by remote sensing and deep learning technologies - a tiered approach

**Nadiia Kopiika[a,g], Andreas Karavias[b,g], Pavlos Krassakis[b,g], Zehao Ye[a], Jelena Ninic[a], Nataliya Shakhovska[e,c,d,g], Nikolaos Koukouzas[b], Sotirios Argyroudis[e,f,g], Stergios-Aristoteles Mitoulis[* a,f,g]**

(a) University of Birmingham, Edgbaston, Birmingham, B15 2TT, UK

(b) Centre for Research and Technology Hellas (CERTH), 15125 Athens, Greece

(c) Lviv National Polytechnic University, Stepana Bandery St, 12, Lviv, Lvivska oblast, 79000, Ukraine

(d) University of Agriculture in Krakow, Podłużna 3, 30-239, Poland

(e) Brunel University London, Kingston Lane, Uxbridge Middlesex, UB8 3PH, UK

(f) MetaInfrastructure.org, London, UK

(g) bridgeUkraine.org, London, UK

*Corresponding author: Stergios-Aristoteles Mitoulis, S.A.Mitoulis@bham.ac.uk

## Abstract

Critical infrastructure, such as transport networks and bridges, are systematically targeted during wars and suffer damage during extensive natural disasters. The former is because critical infrastructure is vital for enabling connectivity and transportation of people and goods, and hence, underpins national and international economic growth. Mass destruction of transport assets, in conjunction with minimal or no accessibility in the wake of natural and anthropogenic disasters, prevents us from delivering rapid recovery and adaptation. As a result, systemic operability is drastically reduced, leading to low levels of resilience. Thus, there is a need for rapid assessment of its condition to allow for informed decision-making for restoration prioritisation. A solution to this challenge is to use technology that enables stand-off observations. Nevertheless, no methods exist for automated characterisation of damage at multiple scales, i.e. regional (e.g., network), asset (e.g., bridges), and structural (e.g., road pavement) scales. Moreover, there is no systematic correlation between infrastructure damage assessments across these scales due to the lack of integrated approaches. We propose a methodology based on an integrated, multi-scale tiered approach to fill this capability gap. In doing so, we demonstrate how automated damage characterisation can be enabled by fit-for-purpose digital technologies. Next, the methodology is applied and validated to a case study in Ukraine that includes 17 bridges, damaged by human targeted interventions. From regional to component scale, we deploy technology to integrate assessments using Sentinel-1 SAR images, crowdsourced information, and high-resolution images for deep learning to facilitate automatic damage detection and characterisation. For the first time, the interferometric coherence difference and semantic segmentation of images were deployed in a tiered multi-scale approach to improve the reliability of damage characterisations at different scales. This integrated methodology automates and hence accelerates decision-making to facilitate more efficient restoration and adaptation efforts, ultimately building resilience into our infrastructure.

## 1. Introduction

Critical infrastructure assets, like bridges, have a vital role for the transportation of goods, accessibility, and the economy, facilitating the flow of people, vehicles, and goods over obstacles like water bodies and valleys. Therefore, systematic maintenance and monitoring of their condition is of paramount importance to ensure their undisrupted operation [1]. During periods of warfare and conflict, bridges are often targeted due to their pivotal role, making them susceptible to frequent attacks. Bridge damage of different extent, significantly impact their functionality and restoration costs, including direct and indirect expenses, such as repairs, replacements, and rehabilitation efforts.[2] Also, damage affects the structural integrity and their load-bearing capacity, often leading to traffic restrictions or closures to prevent safety risks, crucial for regional infrastructure efficiency. Conducting thorough preliminary assessments of damage in inaccessible regions [3],[4] facilitates remote restoration planning and informed decision-making. The presence of violence and threats in conflict zones challenges conventional and traditional approaches to damage assessment, which typically rely on manual detection and on-site surveys [5]. In the event of mass destruction of critical infrastructure followed by limited accessibility, our ability to enhance resilience through rapid assessments and restoration is hindered [6], leading to significant direct and indirect losses, and hence, delays in restoring normal economic activity [7]. However, damage assessment can be facilitated substantially by automated integration of digital technologies [4]. Although individual digital technologies and data sources have been widely adopted, integrating different technologies is seen as a practical approach to addressing gaps that arise from using a single method. Furthermore, there is increasing momentum in data fusion of various methods, scales, and precisions, which allows for the development of more sophisticated and automated data-driven decision making. Various automatic data integration procedures and synergistic combination of data with different spatial and temporal resolutions will facilitate automatic multi-scale exchange of information obtained from disparate sources.[8] Therefore, an integrated assessment framework is needed, combining different scales, i.e. regional, infrastructure asset, and component. Leveraging digital technologies in automatic way, this framework aids in restoration strategies, providing intelligence to decision-makers, governments, and funders for effective investment prioritisation in rebuilding conflict-devastated urban environments.

## 2. State of the art and background

### 2.1. Stand-off observations for damage characterisation

Critical asset safety is typically assessed by periodical site inspections and testing [9],[10], to inform decisions for targeted maintenance, which are typically time-consuming, costly [11], risky and possibly inaccurate [12]. In some cases, manned inspections are impossible, such as in war zones, due to safety risks and inaccessibility. Extensive damage often requires large-scale spatial inspections, which may slow down the recovery process as they rarely account for the importance and interdependencies of assets at the regional scale. This inadequacy renders them insufficient for the effective post-disaster management of large portfolios of assets and regions [13],[14]. Therefore, there is an urgent need for more reliable and rapid decision-making for prioritisation of restoration strategies that will use disparate digital or traditional data sources, available after natural hazards, such as floods [15], earthquakes [16], landslides [17], and conflicts [18] to accelerate recovery [19].

Available methods for infrastructure assessment include e.g., Global Position System (GPS), terrestrial Synthetic Aperture Radar (SAR) Interferometry (InSAR), Internet of Things (IoT) and digital image correlation (DIC) [19],[20],[21],[22],[23]. However, these methods are either effective at the macroscale, facilitating recovery of regions, or at the microscale, e.g., UAVs, GNSS [24],[25], where data is used to develop models of individual assets [26]. More recently, satellite imagery has become a prospective tool for remote evaluation of infrastructure damage. Yet, these technologies have not been integrated in a way to facilitate assessments and decisions at different scales, which is a capability needed to enable efficient restoration strategies.[27] For example, in the case of bridges, stand-off damage characterisation is so far mainly focused at asset or component damage indicators, measuring structural deflections [28],[29], soil settlements [30], cracking [31],[32] and corrosion [33]. Structural health monitoring at asset and component scale with the use of computer vision-based [34] and remote sensing technologies enable the assessment, management, and maintenance of bridges [34], [35],[36]. This way, efficient decision-making toward restoration measures and infrastructure recovery is underpinned both at macro and micro scale. The current advances on the use of InSAR imagery in infrastructure damage assessment is discussed in section 2.2.

### 2.2. Use of InSAR imagery and open data in infrastructure assessment

Earth Observation (EO) technologies are deployed for non-invasive observation and evaluation of affected areas, using satellite images and geospatial data. EO and especially Synthetic-Aperture Radar (SAR) images are increasingly being used as a tool for rapid mapping and damage characterisation after disasters, such as earthquakes and floods [8], [37],[38],[39],[40],[41], [42]. Combination of Geospatial Intelligence (GEOINT) [43],[44], and EO products of geospatial data also facilitates the identification of spatial patterns related to hazard susceptibilities that may lead to infrastructure vulnerabilities [45],[46],[47],[48]. An example is the identification of earthquake-induced building damage using backscatter intensity and phase signals from Interferometric Synthetic-Aperture Radar (InSAR) images [49],[50],[51]. In this case, the Coherent Change Detection (CCD) techniques can utilise the phase signal correlation of InSAR products by comparing land changes before and after the events, to detect the affected areas, enabling, to a certain degree, the characterisation of the damage of infrastructure and the natural environment [51],[52],[53].

Most of the studies thus far (e.g., [36]-[41],[45],[49],[50]) primarily focused on assessing damage as a result of natural hazards. However, structural destructions, caused by human-induced hazards, e.g., wars and, terror attacks, follow different patterns and have different characteristics; damaged assets, are sparsely distributed in intricate urban environments, occupying only a minor part of urban areas, while the majority of the surrounding environment might be unaffected. Such cases are associated with a notable imbalance between damaged and undamaged structures, which is unique in conflict zones but not in areas affected by climate hazards, e.g., floods or earthquakes. As a result, the considerable class imbalance in conjunction with the heterogeneous urban environments cause significant challenges in identifying damaged assets, which makes the damage characterisation very challenging. Such an obstacle could be eliminated by the implementation of additional investigations using open data, e.g., crowdsourcing, Open Street Maps, online open platforms. Another common feature of the majority of studies is their reliance on high-resolution satellite imagery [19]. Unavailability of such high-resolution satellite data during and after the conflict, due to confidentiality and national security, comprise significant hurdles to assessing the damage by this approach. Given the above circumstances, the application of low-resolution satellite imagery, especially in conditions of high heterogeneity of damage level, becomes compelling and noteworthy for the damage characterisation of infrastructure assets on conflict-torn territories.

### 2.3. AI techniques and crowdsourcing in damage detection

Deep learning and Computer Vision (CV) are subset methods of AI focused on the automatic extraction of useful information from image or video data to facilitate the assessment and understanding of the underlying physical world [57]. These technologies have been increasingly used in civil engineering to automatically perform a number of tasks related to inspections, monitoring, and assessment of infrastructure, often complementing, or even replacing manual analysis [58].

Leveraging Machine Learning (ML) and CV methods, we can learn intricate patterns from vast datasets, enabling automatic, highly accurate and efficient damage detection [57],[59]. The recent increase in computation power enhanced the usage of deep learning and Computer Vision for handling a variety of ML tasks in practical scenarios [59],[60],[61].

Possible approaches to localise damage in images include object detection and segmentation. For example, CNNs have been used to classify concrete cracks and determine the types of road damage [62],[63]. However, existing methods often treat damage as a high-level concept rather than a well-defined object, creating conceptual mismatches [64]. To overcome this limitation, we propose damage detection based on heat mapping and Grad-CAM localisation.

Current research on disaster damage detection and assessment relies heavily on macro-level imagery, such as remote sensing imagery [54] or images collected by unmanned aerial vehicles (UAV) [55]. With the growth of social media platforms, real-time information about infrastructure damage and destruction can be found through textual data analysis and images posted by eyewitnesses. [65]. Social media image analysis using convolutional neural networks (CNNs) serves as an auxiliary source for assessing infrastructure damage [56].

Automated vision-based structural inspection using semantic segmentation algorithms enables rapid analysis of the conditions of the infrastructure assets affected by hazards, in conditions of limited time, accessibility and resource constraints [66], [67]. CV tasks enhanced by deep learning empower machines to autonomously discern and identify asset component and the fundamental characteristics of a damaged asset [68]. This involves the utilisation of advanced neural network architectures to automatically extract intricate patterns and relevant features from the provided data. The integration of deep learning and CV tasks significantly enhances the machine's ability to comprehend and interpret complex visual and textual information related to asset damage. This approach includes the following tasks, as described in Figure 1 [69]: (a) image classification based on labelled image, e.g., spalling, crack; (b) patch-wise classification where each patch is classified as either presenting a crack or not; (c) object localisation where bounding box indicates the position of the defect; (d) object localisation based on heatmap; (e) object detection; (f) semantic segmentation to classify individual pixels. Thus, CV algorithms assist in localising and quantifying structural defects and damages [70],[71], eliminating the necessity for labour-intensive and highly subjective on-site inspections [72], [73].

The pre-processing of images for damage detection can been performed based on state-of-the art models like the Segment Anything Model (SAM), capable to generate high-quality masks. Its core involves the establishment of a data engine, which comprises three stages: assisted-manual annotation, semi-automatic annotation, and fully automatic annotation, with refinements and improvements at each stage of the process.

Researchers focus on refining deep learning-based model architecture [74] and enhancing training data quality and quantity [75],[76] to improve damage detection techniques. The Bidirectional Feature Pyramid Network model [77] has been instrumental in locating damage in locating damage with high accuracy (96%). Data augmentation techniques and transfer learning of trained models based on the ImageNet dataset [78] also contribute to improving detection accuracy. A comprehensive comparative analysis between Mask R-CNN and YOLO (You Only Look Once) model is presented in [79]. The study takes into account diverse data types, encompassing visual images, point cloud, infrared thermal imaging, ground-penetrating radar, vibration response, and other relevant types of data. Deep learning methods, thus, offer robust tools for analysing different data sources, promising enhanced efficiency, accuracy, and automation in structural health monitoring.

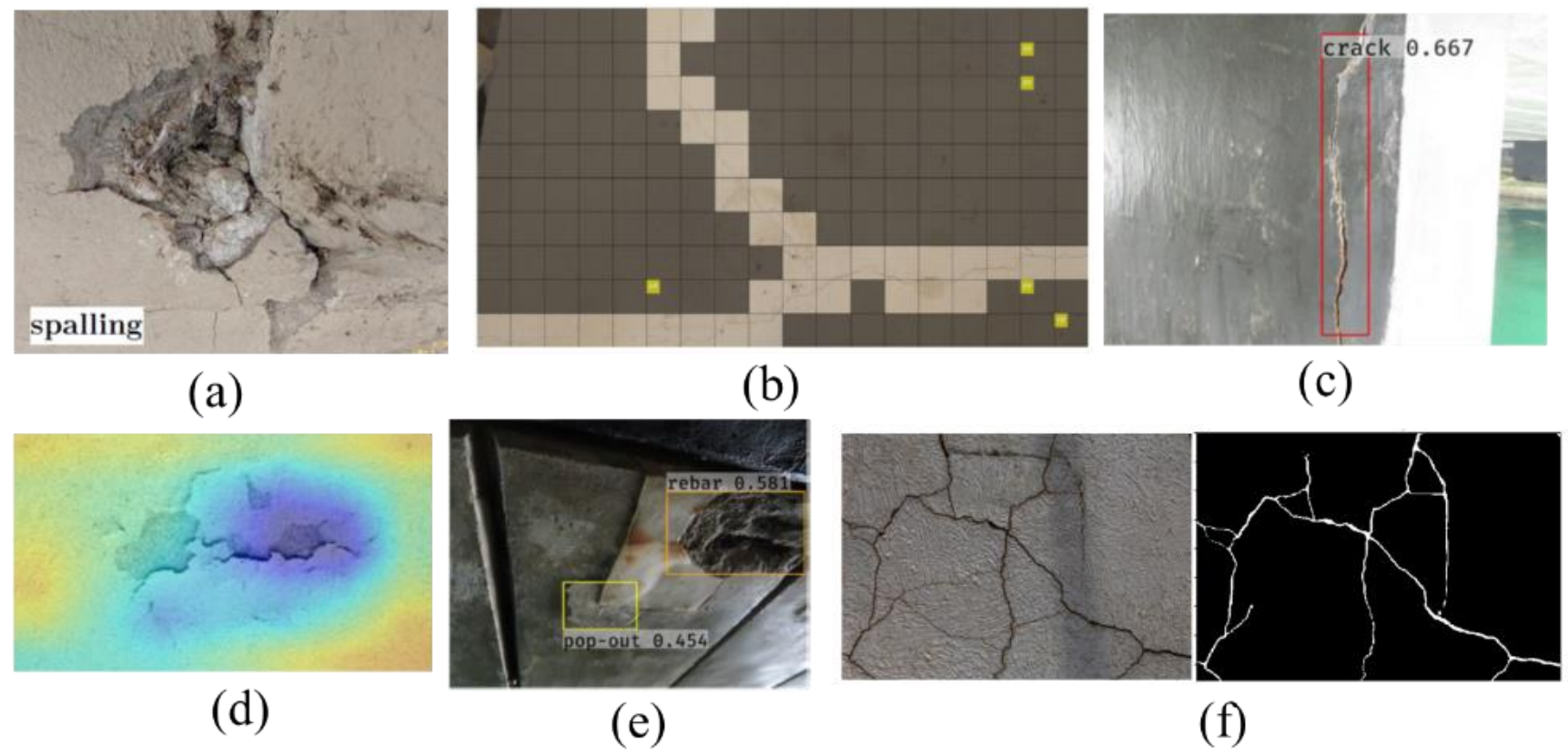

*Figure 1. Computer Vision tasks empowered by Deep Learning for (a) image classification; (b) patch-wise classification; (c-d) object localisation; (e) object detection; (f) semantic segmentation.*

### 2.4. Extensive damage challenges traditional methods - knowledge gaps and novelty

Extensive destruction of bridges, coupled with limited or no access to these critical assets during and on the aftermath of extensive natural or human-induced disasters, hinders our ability to characterise the damage and build resilience into critical infrastructure and communities [6]. This is because the damage assessment that includes (a) the damage mode and cause, (b) the extent, also known as damage level, (c) the accessibility and (d) its interdependencies with other assets and systems, and (e) the availability of resources (funds, labour, materials), are the dominant factors in decision making. These factors shape the adaptation and restoration strategies, and this gave the motivation for this paper which focuses on automatic damage characterisation approach.

Even though there has been extensive research in specific technologies for damage identification at (i) macro (regional) and (ii) meso (asset) and (iii) micro (component) scale, there is no framework that integrates different scales of damage characterisation. Thus, research outcomes have not been integrated into unified engineering framework, while most of research concerns endeavours by computer and/or earth observation scientists, who neglect engineering principles and practice. Therefore, research either misses the importance of asset damage state in (i), which may affect the operability of the region, or misses the state of the functionality level of the region in (ii) that may prevent the timely restoration of assets. In both cases, resilience is dramatically affected by the absence of knowledge integration in damage characterisation at different scales. This is a challenging gap in the knowledge toward which this paper contributes. For example, natural hazard-induced damage and assessment by satellite imagery may not be straightforward for human-induced hazards, e.g., wars, terror attacks, because of the high-class imbalance due to destruction, affecting only a small part of urban areas, surrounded by unaffected environment. The absence of geographical patterns, typical for natural damages, and diverse characteristics of urban environments cause certain challenges in the identification of affected assets within the whole infrastructural system. Moreover, the unavailability of high-resolution satellite data in conflict-prone regions due to security and confidentiality emphasises the unique nature of human-induced hazards.

To the authors' best knowledge, this is the first tiered approach, that integrates disparate open-access sources toward a multi-scale rapid automatic damage characterisation of critical infrastructure in conflict-prone regions. This paper puts forward a framework for the use of disparate technologies and openly available data to characterise damage at different scales from regional, to asset, to component and ultimately enables rapid and well-informed decisions toward restoration (see Figure 2). The focus is on regional networks that include bridges, affected by shelling. This framework identifies damage of critical infrastructure using the InSAR Coherent Change Detection (CCD) method, utilising interferometric coherence difference values to evaluate the damage level. CCD-based assessments are then validated using stand-off observations, e.g., openly available satellite images and photographs as well as inspection records, open data, and crowdsourcing. When asset-specific CCD information is not adequate to make a decision regarding damage characterisation, an asset-scale approach is deployed for damage characterisation. The latter detects damage at component scale, using semantic segmentation for automatic localisation and damage classification. The method scouts a number of appropriate AI pre-trained big models for component-specific damage detection in the context of post-disaster inspection, taking into consideration the uncertainties in the obstruction of the subject and complex backgrounds.

This integrated framework is demonstrated and validated for a case study region in Ukraine, aimed at quickly characterising post-conflict damage in transport infrastructure at asset and component scale within a short time frame. Irrespective of the scale, the framework integrates these methods to conclude with reliable identification of damage level.

## 3. Methodology for rapid multi-scale damage assessment

### 3.1. Methodology, framework, and data for damage analysis at different scales

Figure 2 describes the integrated framework for automated damage characterisation toward decisions for restoration. The figure indicates different scales of assessment, i.e. region (R), asset (A) and component (C). More detail regarding the research specific terminology is available in the Appendix A. The methodology commences with the threat identification and proceeds with the selection of the assessment scale: for all the assets residing within the region of interest, openly available data are used to map the critical assets and their interdependencies. Damage detection at regional and asset scale is conducted (see more in Figure 3), by employing remote sensing technologies, e.g., ESA Open Hub. If the damage characterisation results to high level of confidence, and hence, to accurate damage characterisation, then the only hurdle to designing and applying a restoration strategy is the connectivity, and therefore, accessibility to the asset. For example, if the asset is a bridge, connectivity to the bridge will be sought both through the road network, which the bridge belongs to, and through other routes that may lead to critical regions. If access to the asset is possible, then we can proceed with decisions, detailed designs, and the restoration strategy. If not, then the level of damage of the connection (e.g., road connecting the bridge) should be assessed using the same method. In this case, damage characterisation for the connectivity (e.g., roads) is also conducted by remote sensing.

If damage characterisation at scale (A) is insufficient for making decisions and designs for restoration, then additional information is required to inform decision-making. That would lead to higher level of accuracy by characterising damage at (C) scale, using high-resolution images from open-access platforms (see Figure 4). Based on this a decision can be made that the damage characterisation is adequate to proceed with the restoration strategy. Otherwise, better quality data should be sought (e.g., testing, inspections) to proceed with designs of restoration and adaptation measures.

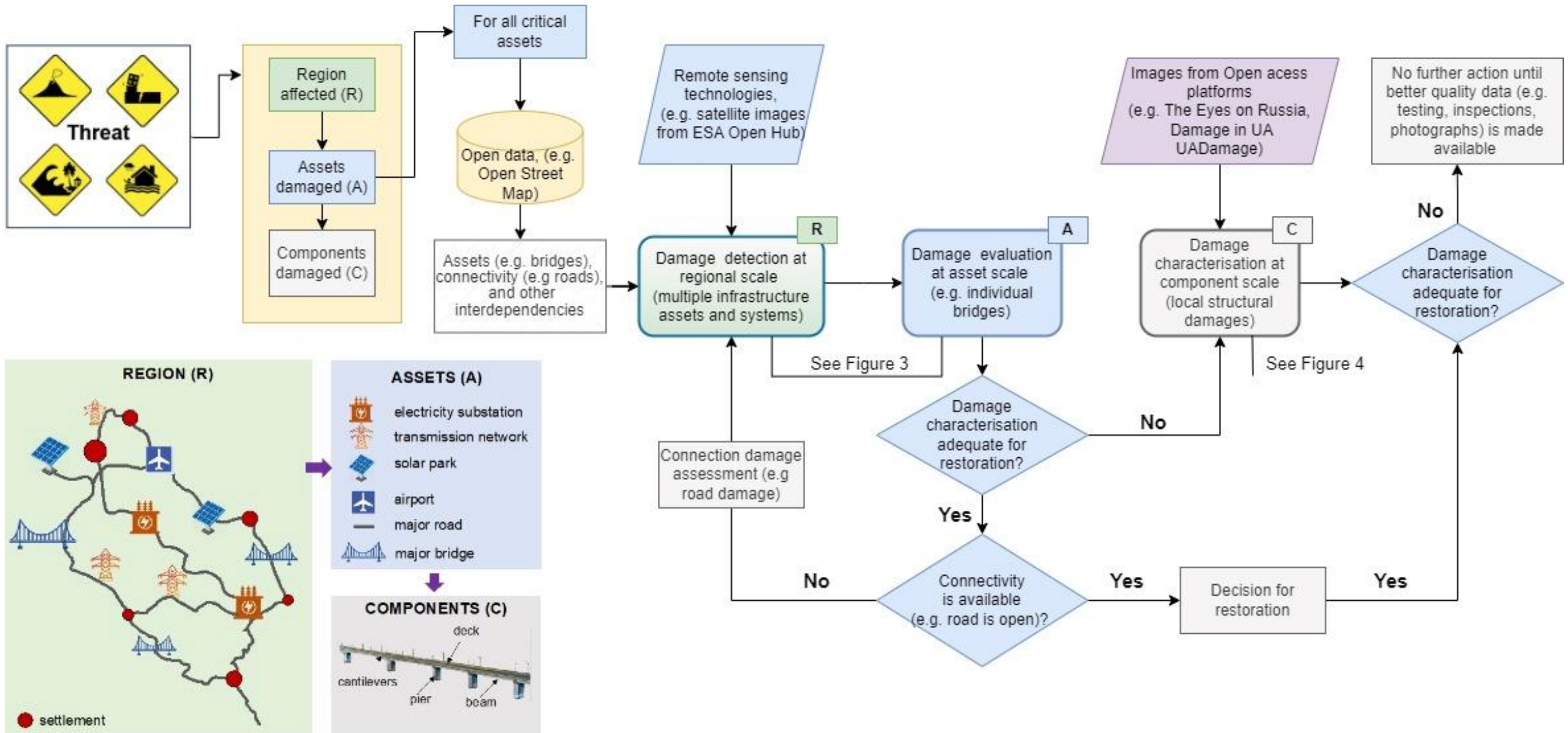

*Figure 2. Framework for automated damage characterisation at different scales: regional (R), asset (A) and component (C) toward decisions for restoration. The framework description continues with figures 3 and 6.*

### 3.2. Method for damage characterisation at regional and asset scale

Damage characterisation at regional and asset scale comprises utilisation of open-access satellite imagery, such as Sentinel-1 Single Look Complex (SLC) products and crowdsourced data, e.g., OpenStreetMap (OSM) data for the period of interest, i.e., the time during which extensive damage is inflicted at the area of interest. The damage evaluation at regional and asset scale is performed with the four phases described in Figure 3. A more detailed version of this flowchart is available in the Appendix A.

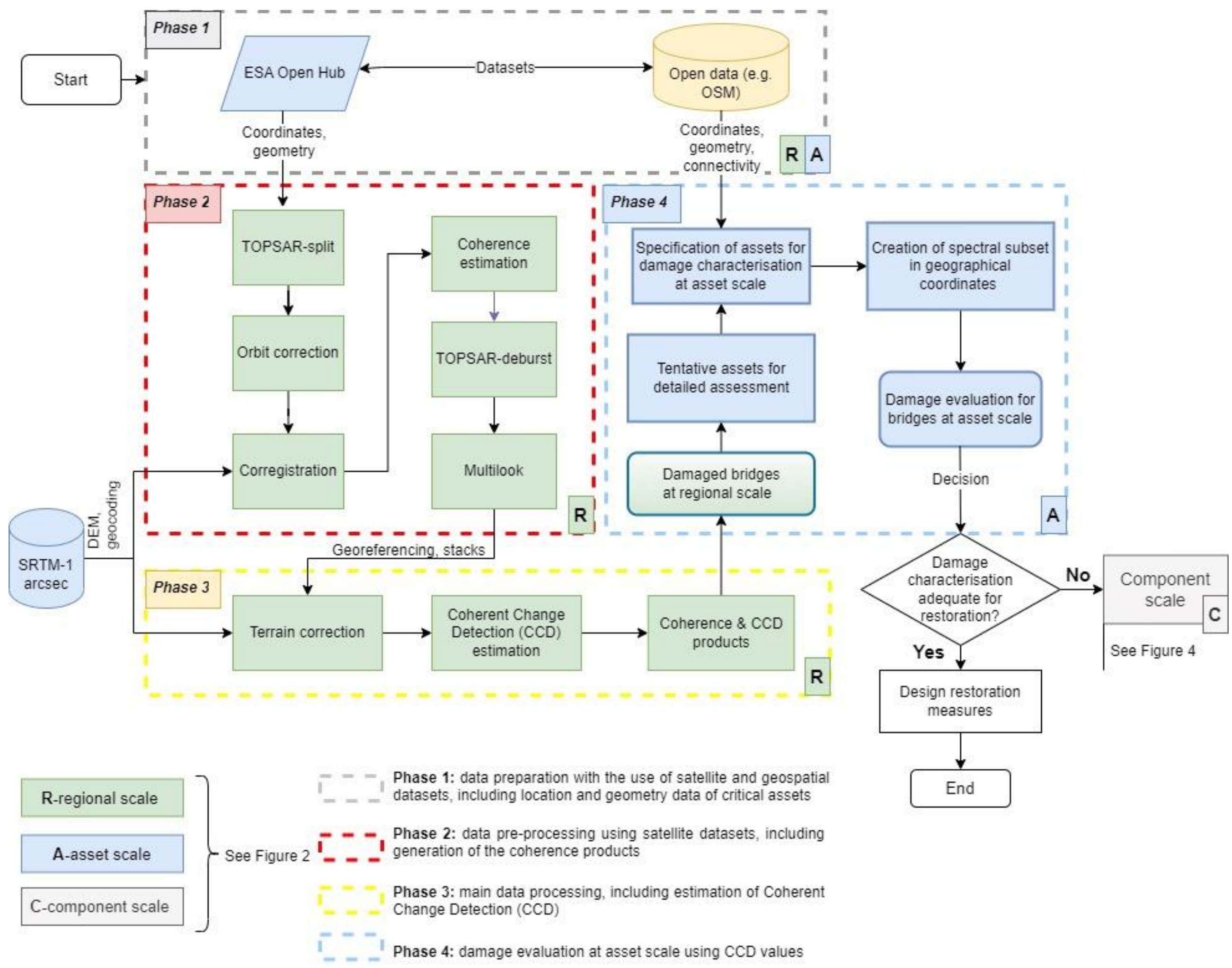

*Figure 3. Workflow for damage evaluation at regional (R) and asset (A) scale based on four phases: Phase 1 (grey-dashed box): data preparation with the use of satellite and geospatial datasets, including location and geometry data of critical assets, residing within the boundaries of the selected study area; Phase 2 (red-dashed box): data pre-processing using satellite datasets, including generation of the Coherence products (e.g., Sentinel-1 SAR SLC images); Phase 3(yellow-dashed box): main data processing, including estimation of Coherent Change Detection (CCD) and development of a semi-automated method for the damage detection on infrastructure assets, e.g., bridges; and Phase 4 (blue-dashed box): damage evaluation at asset scale using CCD values.*

In **Phase 1,** the Sentinel-1 mission interferometric wide swath (IW) SAR images are obtained in ascending and descending geometry, covering the region of interest, and the assumed time of hazard from open access platforms (e.g., ESA Open Hub) [80], [81]. Simultaneously, for the preliminary estimation of the geographic coordinates of the affected assets, data can be obtained from open-access data platforms, e.g., OpenStreetMap (OSM) [83] and crowdsourcing [84],[85],[86]). This includes, e.g., transport route disruption, disrupted connectivity in the region, and destruction of assets and connections reported in social and other open-source platforms. **Phase 2** includes the pre-processing of Sentinal-1 SAR SLC images using the Sentinel Application Platform (SNAP) architecture [82]. The left part of the workflow describes the process of splitting the images on the selected sub-swath with the specific bursts that cover the study area (TOPSAR split). This includes orbital correction by using the precise orbit files (Orbit correction). Subsequently, the images are corregistrated using the Digital Elevation Model of Shuttle Radar Topography Mission (DEM SRTM) -1 arcsec [87]. With this exercise the interferometric pairs before and after the damage are generated. A coherence estimation is implemented in every interferometric pair followed by the "TOPSAR-deburst" and "Multilook" steps. Next, the coherence products are geocoded using the DEM SRTM-1 arcsec and grouped in stacks in order to calculate the CCD. The damage detection takes place during **Phase 3,** the main processing stage, which involves the geospatial analysis for two stacks of images: a pair of pre-damage products and a pair of one pre- and one post- damage products. First, the image pairs are georeferenced via Terrain correction. Then, InSAR coherence products, calculated for the two pairs of SAR images, are calculated, serving as an indicator of the similarity in radar reflections between examined datasets. The result represents the level of correlation in the phase of the corresponding pixels of the two images. Any changes in the backscattered signal of the satellite are recognised as decorrelation of the phase. As a result, changes in the scene from one acquisition to the next are detected and recorded. Changes between the two images reduces the coherence value and negatively affects the accuracy over the distance measurement between the antenna of the satellite and Earth's surface [88]. The coherence (γ), which is also defined as the complex correlation coefficient between two SAR scenes, $u_1$ and $u_2$, is estimated using Equation (1) [89]:

$$\gamma = \frac{E\left[u_1 u_2^*\right]}{\sqrt{E\left[\left|u_1\right|^2\right]}\sqrt{E\left[\left|u_2\right|^2\right]}} \qquad \text{Equation (1)}$$

where *E{}* represents the mathematical expectation and * is the complex conjugate operator. The coherence values range from low (γ=0) to high (γ=1) (high coherence). Pixels with high coherence values are characterised as stable, as they have very small variations over time. Low coherence values indicate significant changes. Coherence is utilised here to identify damage of the built environment [89],[53], in that any change in the visible plan view of the asset would be identified as CCD. Regarding the calculation of CCD, this requires three images: a pair of images that are acquired before the event *(pre)*, and another pair of images, one obtained before and one after the event *(post)* [90], that causes the change of CCD. CCD values range from -1 to 1. Positive values represent areas with significant differences, indicating changes in the region under study that includes the built environment and/or the ground surface. Values close to zero indicate stable areas between satellite passes, while negative values are new stable areas appearing during the interval between the two coherence products. The CCD is calculated as per Equation (2):

$$CCD = \gamma(pre) - \gamma(post) \qquad \text{Equation (2)}$$

For example, if the coherence of the two images acquired before the hazard event is high, e.g., $\gamma(pre)=0.9$, this will show high correlation between the images and high stability in the built environment, i.e., no change or damage occurred during the time that the two images were taken. If the coherence between a pair of SAR images obtained before and after the hazard event is lower, e.g., $\gamma(post)=0.5$, it indicates, that the hazard event has resulted in damage changes in the investigated area. Thus, CCD is a measure of change which is correlated here to the infrastructure damage level.

In **Phase 4**, coherence and CCD products are integrated into a Geographic Information System (GIS), e.g., ArcGIS, QGIS, to illustrate the built environment or ground surface, where CCD values indicate potential change, and hence, damage if this refers to structural assets. Specifically, the products of this phase highlight the changes between periods before- and after- the induced damage, offering a semi-automatic way of detecting significant changes. Based on that, the coherence and CCD products are considered mutually in order to focus on areas close to the investigated asset of interest. Thus, coherence tells us about the consistency or stability of surfaces, and CCD products help us spot changes over time and to pinpoint areas thus identify signs of damage that may require inspection and repair. After defining their coordinates, assets of interest are examined in more detail using open data sources (e.g., Google Maps, OSM) and Setinel-1 images for cross-validation and specification for the final assessment. The spectral subset region in geographical coordinates using WKT-format is used to indicate the area of interest of each assessed asset. Then, the results are exported to GIS environment for illustration, geographical collocation, and damage characterisation at asset scale. Thus, different ranges of CCD results indicate different damage levels, which can be local or global. This classification of damage to different prescribed levels can be performed based on engineering criteria as described in section 4.2 and Figure 10.

### 3.3. Method for automated damage characterisation at component scale

When the regional or asset scale assessment is not adequate to make decisions regarding restoration, a detailed component-scale assessment is required. This includes detection and automatic localisation and classification of damage using semantic segmentation, which is described from a methodological point of view and also illustrated in [91],[92]. For assets, detected with high coherence values for which macroscopic remote sensing data is not adequate to make decision, further visual information is collected from open platforms. For example, in the case study of this paper, images from Damage In UA and UADamage have been used (see section 4). The images that have visible structural damage are then processed using selected CV techniques to automatically detect the condition of the components. For this, two steps are required: (i) component segmentation, for the detection of specific components of the structure such as deck or pier of a bridge; and (ii) instance segmentation, for the assessment of damage types for the component. As in some instances, images are taken under adverse conditions that limit the image resolution, targeted techniques for image pre-processing based on large pre-trained foundational models are used to improve the quality of images and remove the occlusion. Applying these large pre-trained foundational models expanded the capabilities to downstream and customise our CV tasks. Similarly, models pre-trained for image-text matching, like Grounded Language-Image Pre-training (GLIP) [93],[94] and Contrastive Language-Image Pre-Training (CLIP)[95] were used for this purpose. The pre-processing of all images for component damage detection was performed based on state-of-the art model for instance segmentation, the Segment Anything Model (SAM) [91]. SAM model can be seen in Figure 4a. Figure 4b illustrates an SAM-based architecture with a self-generating prompter mechanism.

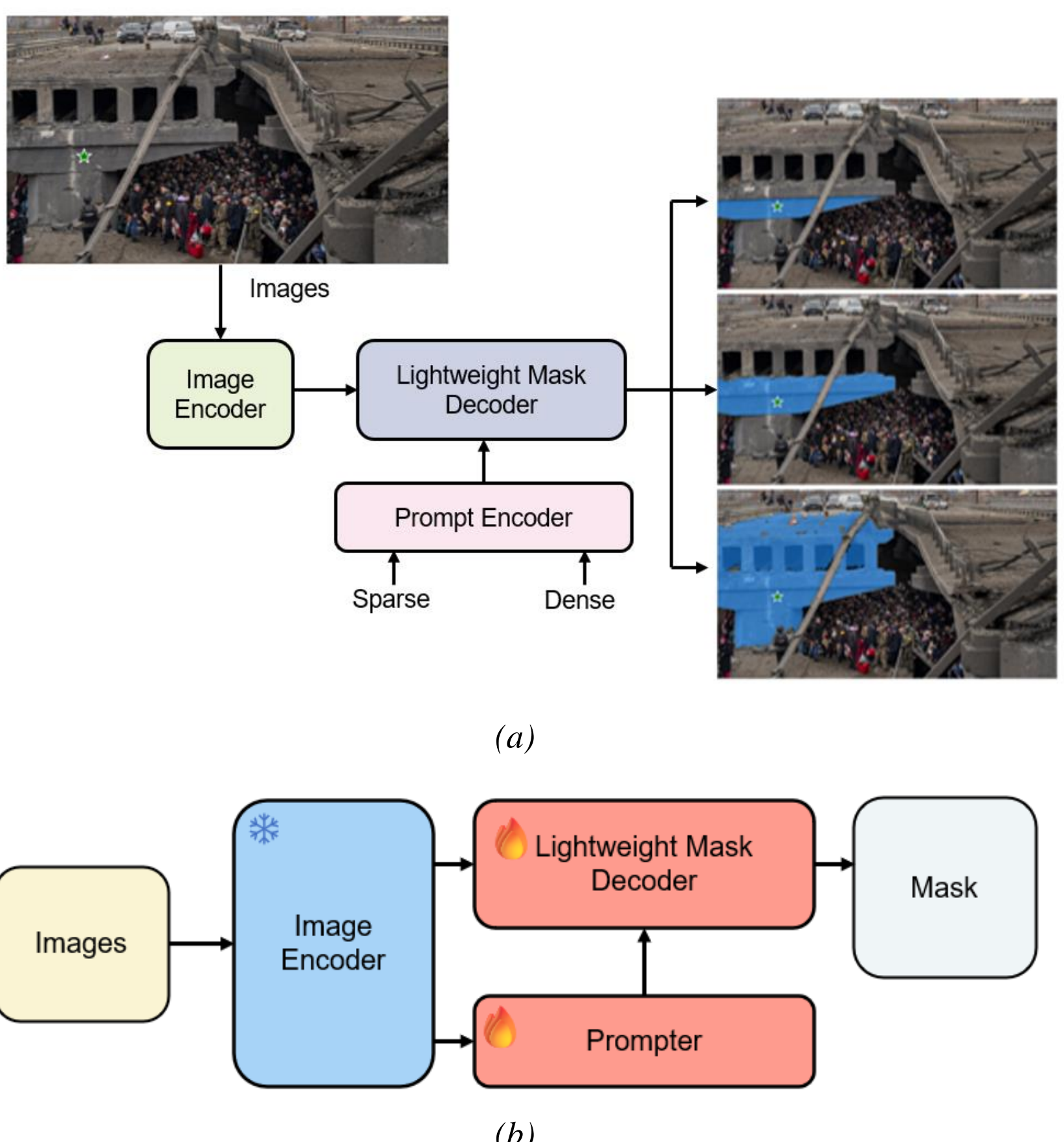

*Figure 4.* ***(a) Segment Anything Model Overview*** *[91]: The architecture of the SAM model includes a heavyweight image decoder, a prompt decoder, and a lightweight mask encoder. The image decoder generates image embeddings, and the prompt decoder accepts two types of prompts: sparse (points, bounding boxes, text) and dense (masks) from human, converting prompts into prompt embeddings. The mask decoder generates*

*corresponding masks based on both the image and prompt embeddings. The diagram illustrates an input with a point prompt (green star) on the main body of a bridge. SAM then generates three different masks corresponding to the whole, parts, and subparts of the bridge. Such original SAM architecture relies on manual prompts and cannot autonomously perform CV tasks;* ***(b) SAM-based RSPrompter with prompter*** *[92]: This architecture replaces the prompt encoder with prompter which can receive output from the image encoder and extracting key features to train itself in generating the required prompts, enabling the algorithm to execute automatically. Heavy SAM's image encoder will be frozen and not participate in training. Such architecture can perform CV tasks automatically, and maintain state-of-the-art levels compared with other instance segmentation models.*

A suitable prompt can accurately generate the masks required by the user. As shown in Figure 5, only two prompt points are needed to select the most matching mask, which is the crack in the figure that is used for component damage characterisation. This new method transforms the way we assess damage at structural component level.

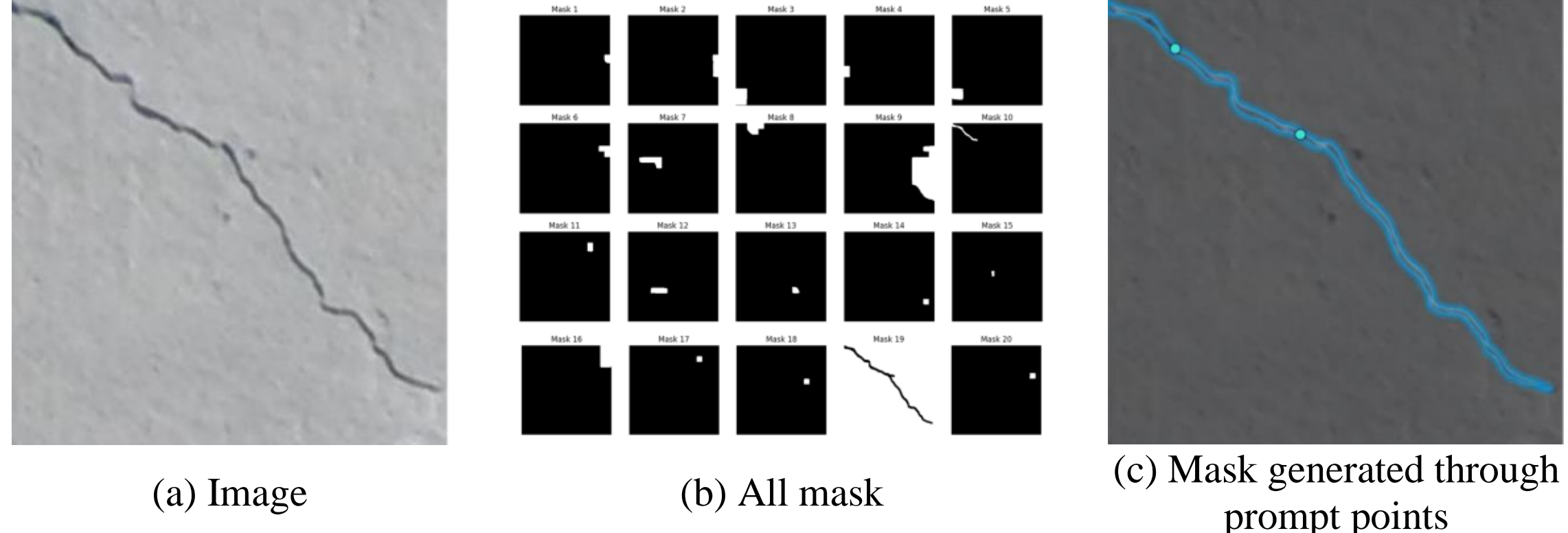

(a) Image (b) All mask (c) Mask generated through prompt points

*Figure 5. Generation of the mask required by the user through prompt points at component scale*

In this paper, various image pre-processing techniques were employed to achieve component automatic damage characterisation, including Grounded-SAM, which integrates Grounding DINO (self-distillation with no labels) [93] and SAM, built upon GLIP [94]. It is designed for open-set object detection and utilises image-text pairs to assign tags to masks generated by SAM. In this paper, we used common open-vocabulary detection, such as the identification of bridges, road, and vehicles. Stable diffusion [96] is a generative AI model based on deep learning, a widely recognised image generation algorithm, which is employed for image inpainting to eliminate occlusions that obscure our primary detected objects, like bridges. By applying an anti-diffusion process on the image, its greatest advantage lies in generating highly relevant and context-consistent repair content. Figure 6 presents our workflow diagram for automatic localisation and classification of damage at component scale.

Starting with the selected images collected from open platforms, the input image is firstly processed using SAM's anything mode to detect all potential masks present in the image. Subsequently, each mask is assigned labels, and those with the "bridge" label are singled out. Following this step, a decision is made based on whether there is occlusion, in which case damage characterisation follows the occlusion restoration (blue dashed box in Figure 6) or not (green dashed box in Figure 6), where damage characterisation is performed.

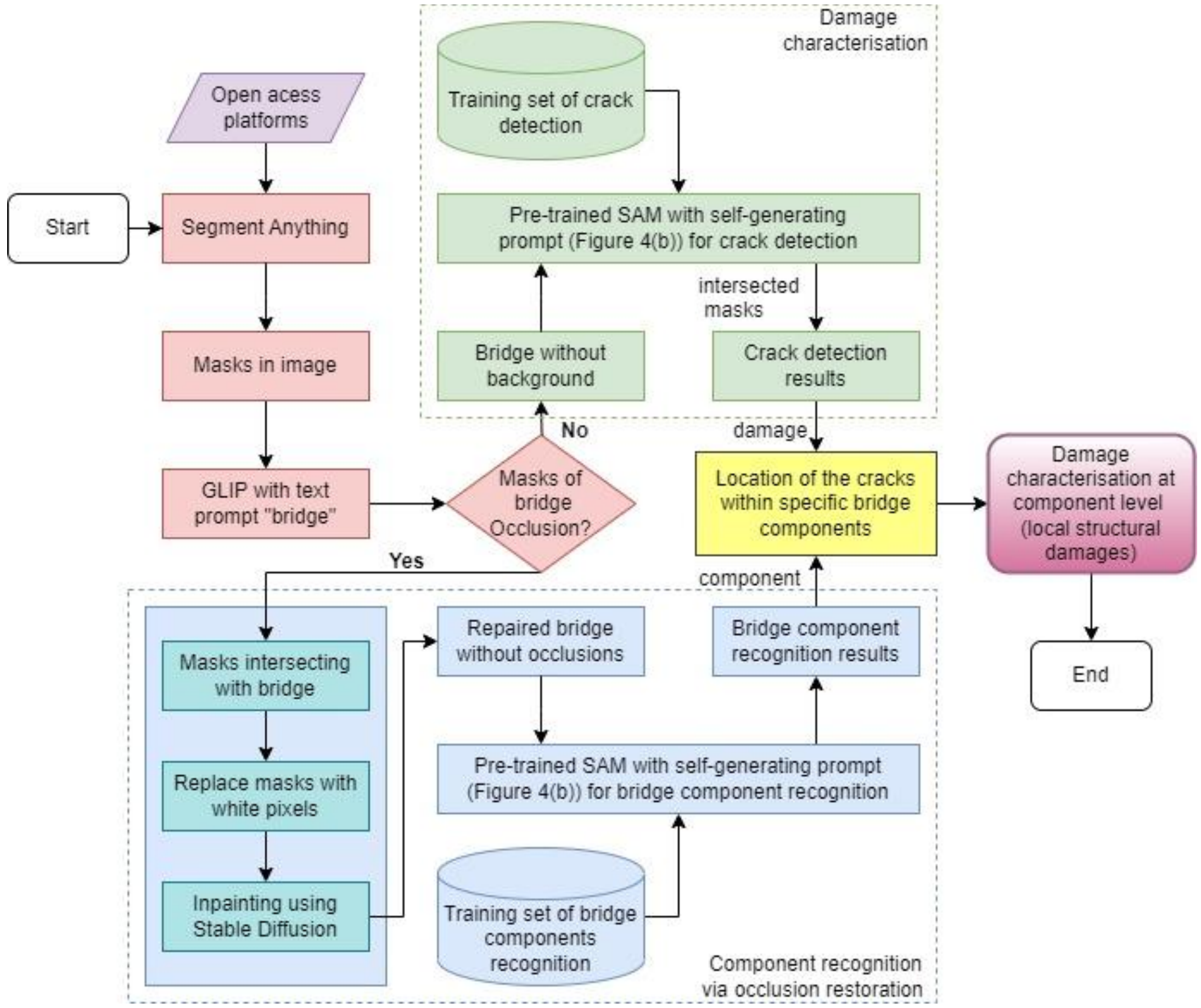

*Figure 6. Flowchart, illustrating the damage characterisation method at component scale.*

The damage characterisation process utilises the masks of bridge identified previously, excluding all pixel except those within the bridge masks. Utilising a pre-trained SAM with self-generating prompt, the model can automatically detect damage, such as cracks. The other process first captures masks intersecting with the bridge. These intersected masks are considered areas requiring treatment, employing stable diffusion for inpainting. This process generates the missing parts of the bridge structure due to occlusions. Subsequently, similar to the process of damage characterisation, another pre-trained SAM with self-generating prompt automatically performs bridge component recognition tasks based on the repaired bridge image. Finally, the identification outcomes from both processes merge, enabling the determination of where the damages have occurred within the bridge components. This is a fully automated procedure that facilitates the damage characterisation for detection of structural damages at the component scale.

## 4. Application to a case study: analysis, results, and discussion

### 4.1. Description of the case study area

The case study is an inaccessible region, for which it is challenging to assess infrastructure damage toward restoration measures. In Ukraine, extensive destruction of civil infrastructure has taken place as a result of missile attacks, shelling, and artillery fire. Roads and traffic have been extensively disrupted, due to the damage inflicted on over 345 bridges across the country [97]. This damage was more pronounced in the Kyiv region and in particular the bridges along the Irpin river, leading to systematic damage and disruption of connection routes, e.g., Bucha-Kyiv, Hostomel-Kyiv, Irpin-Kyiv. These bridges are particularly important because they serve logistic and supply routes, and facilitate the evacuation of civilian population of the capital through humanitarian corridors (see Figure 7 [98], [99]). The critical role of the bridges in this region, their considerable damage, and the fact that they are not accessible for assessment and decision-making due to the ongoing hostilities, gave the motivation for this case study (Figure 7). The aim of this case study is to validate the efficiency of the framework proposed herein, by identifying and characterising the level of damage of selected bridges along Irpin river, and therefore, enhance the resilience of the area by accelerating decision-making. The assessment is at the three scales described above (from regional to component), by implementing the methodology described in section 3.

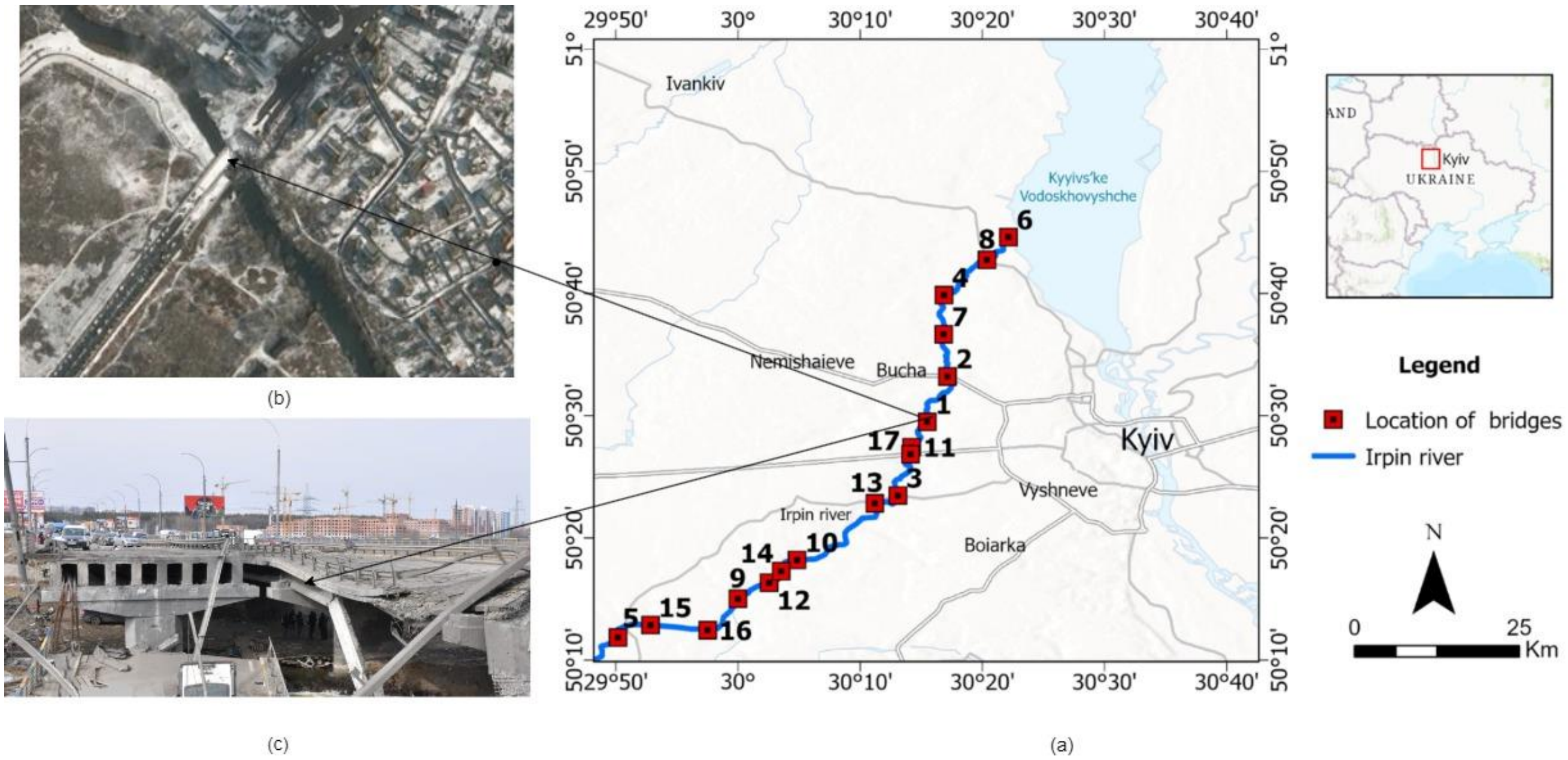

*Figure 7. (a) Case study area, west of Kyiv, Ukraine (the numbers indicate bridge IDs). Examples of damage evidence on bridge over the Irpin river (B1 in this case study) from open-access data: (b) satellite imagery of the damaged bridge captured by Maxar [100] and (c) witness of extensive destruction of the same bridge from social media [101].*

### 4.2. Regional and asset scale damage characterisation

The methodology described in section 3.2 is implemented initially at **regional scale**. Open geospatial data from OpenStreetMap are utilised to select the river crossing bridges essential to connectivity. Alternative routes are sparse or unavailable and hence the network is of low redundancy. This observation was used as a selection criterion and hence bridges that can be bypassed through alternative routes, thus, are less critical for connectivity, were not included in this analysis.

Bridges crossing the Irpin river in the area of interest were visually validated and geometrically corrected using open access Google Satellite Imagery and high spatial resolution optical images of the Maxar basemap in ArcGIS pro. Initially, a total of 24 assets (ID 1 to 24) were processed, which, according to OpenStreetMap serve as bridges. Then, the coordinates of the assets were automatically identified for further localised analysis as per Table A.1 in Appendix A**.** The same table provides information for the structures analysed, also openly available online on OpenStreetMap and Google Maps, including bridge types and their dimensions. Next, each asset was processed individually in detail. Google Maps and Setinel-1 images were used, while a number of assets were excluded from the list. The ones excluded were the assets with ID 18 to 21, which were confirmed that were not serving as typical bridges. The last filter applied was based on the length of the assets, and hence, assets with ID 22, 23 and 24 were also excluded, due to very small length (<10m) that corresponds to culverts, thus can be bypassed by temporary works. A total of 17 bridges went through filtering for further processing, i.e. assets with ID 1 to 17 (see Table A.1). Following screening, the case study area at the west of Kyiv was delineated, and for this area appropriate radar images were obtained and processed at the asset scale.

At **asset scale**, SAR Single Look Complex (SLC) images of Sentinel-1 mission for the period of interest, i.e. time of human interventions leading to damage of critical infrastructure in the Kyiv region, were used to detect and evaluate damage. To obtain the coherence products, three Sentinel-1 mission interferometric wide swath (IW) SAR images were obtained in ascending and descending geometry covering the time period from February 2022 to March 2022, i.e., the time when extensive destruction of infrastructure occurred. For the analysis, the coherence values for two pairs Sentinel-1 images were estimated, describing two time periods (TP); TP1: 19/01/2022 and 12/02/2022, TP2: 12/02/2022 and 01/04/2022. Typically, as the two images of TP1 were acquired in a short order and under similar conditions, i.e., the same incidence angle and environmental conditions, the coherence, which expresses the similarity of the radar reflection between them is expected to be high, ideally close to 1 However, the method for damage evaluation at regional and asset scales relies primarily on the utilisation of freely available Sentinel-1 products, albeit with certain limitations. Hence, the data obtained from low-resolution satellite imagery is considered here with caution in regard to its accuracy. This is because results are not always appropriate for bridge post-disaster damage characterisation at asset scale, due to low image resolution, inappropriate sensor characteristics, low radar frequency and small size of the asset analysed.

As the use of Setinel-1 low resolution images provides limited opportunity for accurate identification of the damage level, for some of the assets the resolution of images was not sufficient. This is because the coherence values were very low (e.g., below 0.5) for the pairs of the images examined. However, for some structures the proposed approach has demonstrated the outstanding capability of damage assessment when access is restricted. The pairs of images, for which coherence results before the hazard occurrence (TP1) were high (above 0.7), are of high reliability. Taking into account these remarks, the coherence between the first image pair was used to assess the applicability of the method developed in damage characterisation and decision making. All assets were classified by the Level of Knowledge (LoK) that reflects the degree of reliability of results. This is based on the image resolution that influences the coherence between the pair of images at TP1. Three LoK are considered based on engineering judgement, i.e., low ($LoK_L$), medium ($LoK_M$), and high ($LoK_H$). All the coherence values below 0.5 indicate low quality of satellite products for damage evaluation in this paper (see Table 1, Figure 8 and Figure 9).

Two coherence products were analysed in this case study: that are the local ($\gamma_{LOC}$), indicating the maximum coherence between corresponding pixels in the pair of images, and global ($\gamma_{GL}$), which indicates the range, for which 95% of the data is within two standard deviations ($2\sigma$) of the mean value (see more details in Appendix A). Figure 8 shows a general trend of reduction of the coherence values for all bridges examined, irrespective of the damage level and the degree of reliability of results. Table 2 shows the numerical results of the maximum local values of coherence ($\gamma_{LOC}$) for each pair of images (TP1, TP2) as per columns 4 and 5, and values, with lower dispersion from the mean in the analysed area, see columns 2 and 3.

**Table 1.** Data level of knowledge (LoK) and appropriateness for damage characterisation

| LoK | $\gamma_{LOC}$ | $\gamma_{GL}$ | Colour legend | Reliability level of results | Decision making/ Usefulness |
|---|---|---|---|---|---|
| $LoK_H$ | 0.75-1.0 | 0.70-1.0 | (green) | High | Method and open-access data applicable for damage evaluation at asset scale. Appropriate for decision making and damage characterisation at component scale |
| $LoK_M$ | 0.55-0.75 | 0.50-0.70 | (orange) | Medium | Method and open-access data useful for damage evaluation at asset scale on some occasions (e.g., for very substantial damage). Additional information (e.g., inspections, crowdsourcing, images with higher resolution) may be needed for decision making |
| $LoK_L$ | 0.0-0.55 | 0.0-0.50 | (grey) | Low | Method and open-access data not applicable for damage evaluation at asset scale. Additional information, e.g., inspections, is needed for decision-making/restoration |

**Table 2.** Results of post-disaster damage assessment using Sentinel-1 SAR images at asset scale

| Bridge ID | $\gamma_{GL}$ | | $\gamma_{LOC}$ | | CCD | | LoK | DL* |
|---|---|---|---|---|---|---|---|---|
| | Before (TP1) | After (TP2) | Before (TP1) | After (TP2) | $CCD_{GL}$ | $CCD_{LOC}$ | | |
| (1) | (2) | (3) | (4) | (5) | (6) | (7) | (8) | (9) |
| B1 | 0.816 | 0.501 | 0.829 | 0.517 | 0.523 | 0.632 | $LoK_H$ | $DL_H$ |
| B2 | 0.859 | 0.611 | 0.967 | 0.829 | 0.499 | 0.540 | $LoK_H$ | $DL_H$ |
| B3 | 0.625 | 0.437 | 0.651 | 0.461 | 0.375 | 0.384 | $LoK_M$ | $DL_M$ |
| B4 | 0.229 | 0.211 | 0.376 | 0.295 | 0.118 | 0.241 | $LoK_L$ | NA |
| B5 | 0.633 | 0.387 | 0.652 | 0.389 | 0.333 | 0.387 | $LoK_M$ | $DL_M$ |
| B6 | 0.876 | 0.717 | 0.889 | 0.754 | 0.144 | 0.390 | $LoK_H$ | $DL_L$ |
| B7 | 0.567 | 0.527 | 0.570 | 0.558 | 0.142 | 0.156 | $LoK_M$ | $DL_L$ |
| B8 | 0.359 | 0.433 | 0.436 | 0.435 | -0.112 | -0.115 | $LoK_L$ | NA |
| B9 | 0.889 | 0.330 | 0.890 | 0.338 | 0.666 | 0.730 | $LoK_H$ | $DL_H$ |
| B10 | 0.469 | 0.506 | 0.469 | 0.506 | -0.145 | -0.145 | $LoK_L$ | NA |
| B11 | 0.588 | 0.526 | 0.588 | 0.526 | 0.280 | 0.280 | $LoK_L$ | NA |
| B12 | 0.504 | 0.446 | 0.526 | 0.456 | 0.188 | 0.189 | $LoK_L$ | NA |
| B13 | 0.406 | 0.346 | 0.505 | 0.401 | 0.087 | 0.178 | $LoK_L$ | NA |
| B14 | 0.406 | 0.231 | 0.505 | 0.313 | -0.029 | 0.062 | $LoK_L$ | NA |
| B15 | 0.567 | 0.264 | 0.683 | 0.376 | 0.350 | 0.400 | $LoK_M$ | $DL_H$ |
| B16 | 0.549 | 0.204 | 0.567 | 0.208 | 0.249 | 0.259 | $LoK_L$ | NA |
| B17 | 0.821 | 0.647 | 0.941 | 0.756 | 0.351 | 0.521 | $LoK_H$ | $DL_M$ |

**NA (not applicable)- bridges, for which the estimated coherence between SAR images from the time period 1 (TP1) was low were classified as those having Low Level of Knowledge ($LoK_L$) and thus were excluded from further analysis. Three DL (damage levels) were defined for assets, see more details in Table 3, Figures 10-11.*

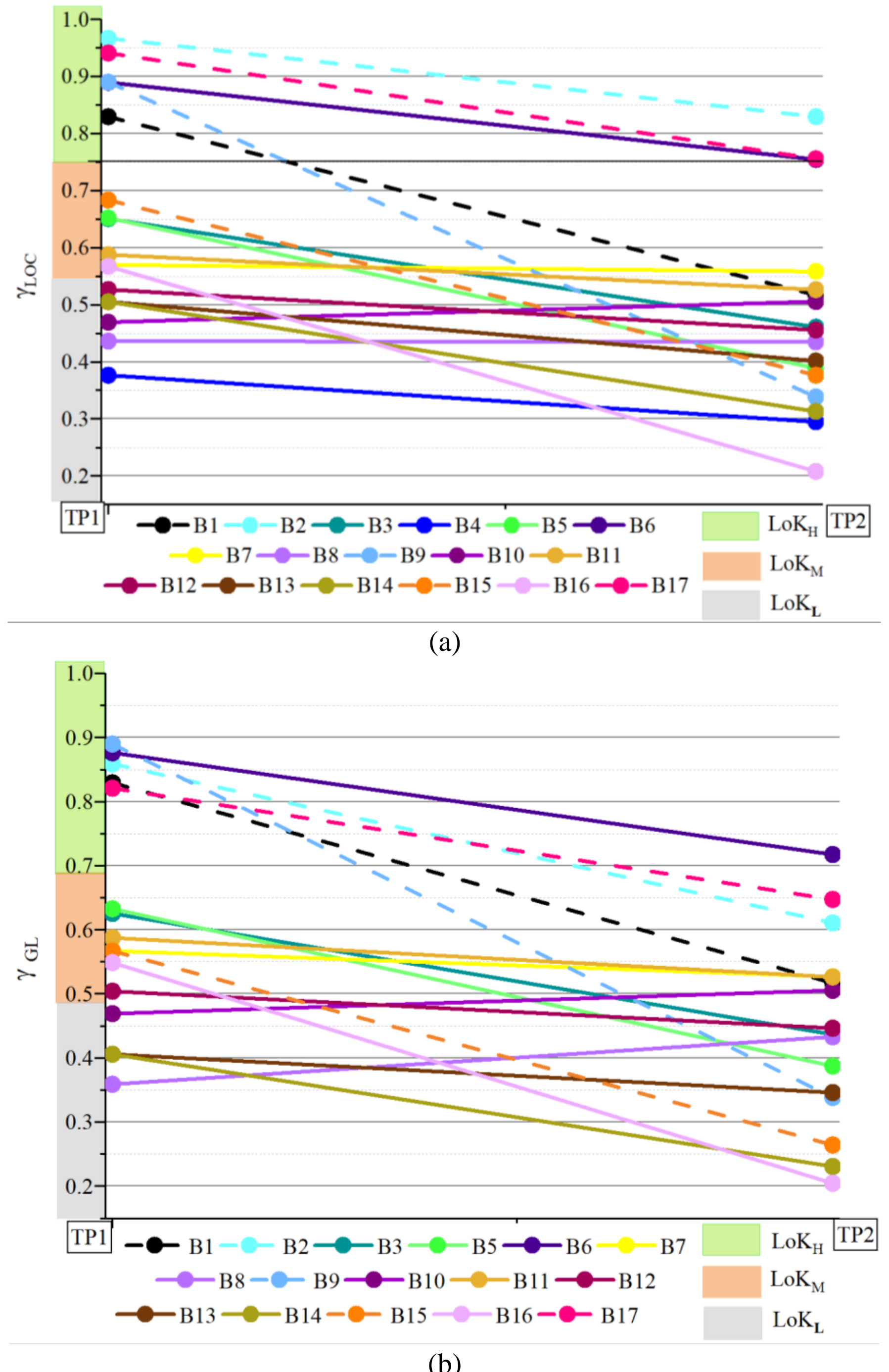

*Figure 8. Change of coherence values within the estimated period (before and after the damage): (a) maximum values for local damage characterisation, (b) mean values, for global damage characterisation. Dashed lines correspond to bridges with greatest coherence changes. Time period 1 (TP1) includes temporal dataset before the beginning of destruction (between 19/01/2022 and 12/02/2022), while time period 2 (TP2), corresponds to dataset of extensive shelling (between 12/02/2022 and 01/04/2022).*

The Coherent Change Detection (CCD) approach is used for post-disaster damage assessment derived from the difference between the coherence products before (TP1) and after the onset of the war (TP2) (see section 3.2). Based on this approach, CCD values were used as the identifier of how much the period of extensive destruction, e.g., missile attacks to infrastructure between February and March 2022, resulted in bridge damage along the Irpin river.

However, as mentioned above, damage detection and evaluation were not possible for all assets of the case study. Thus, the LoK was coined as the criterion to exclude assets for which the information does not provide the required accuracy, see Figures 8 and 9, where different ranges of reliability are clearly indicated. Bridges with values of LoK<0.4, were neglected during damage evaluation, as this data were not reliable for assessing the level of damage (for instance for B4, B8, B10-B14, B16).

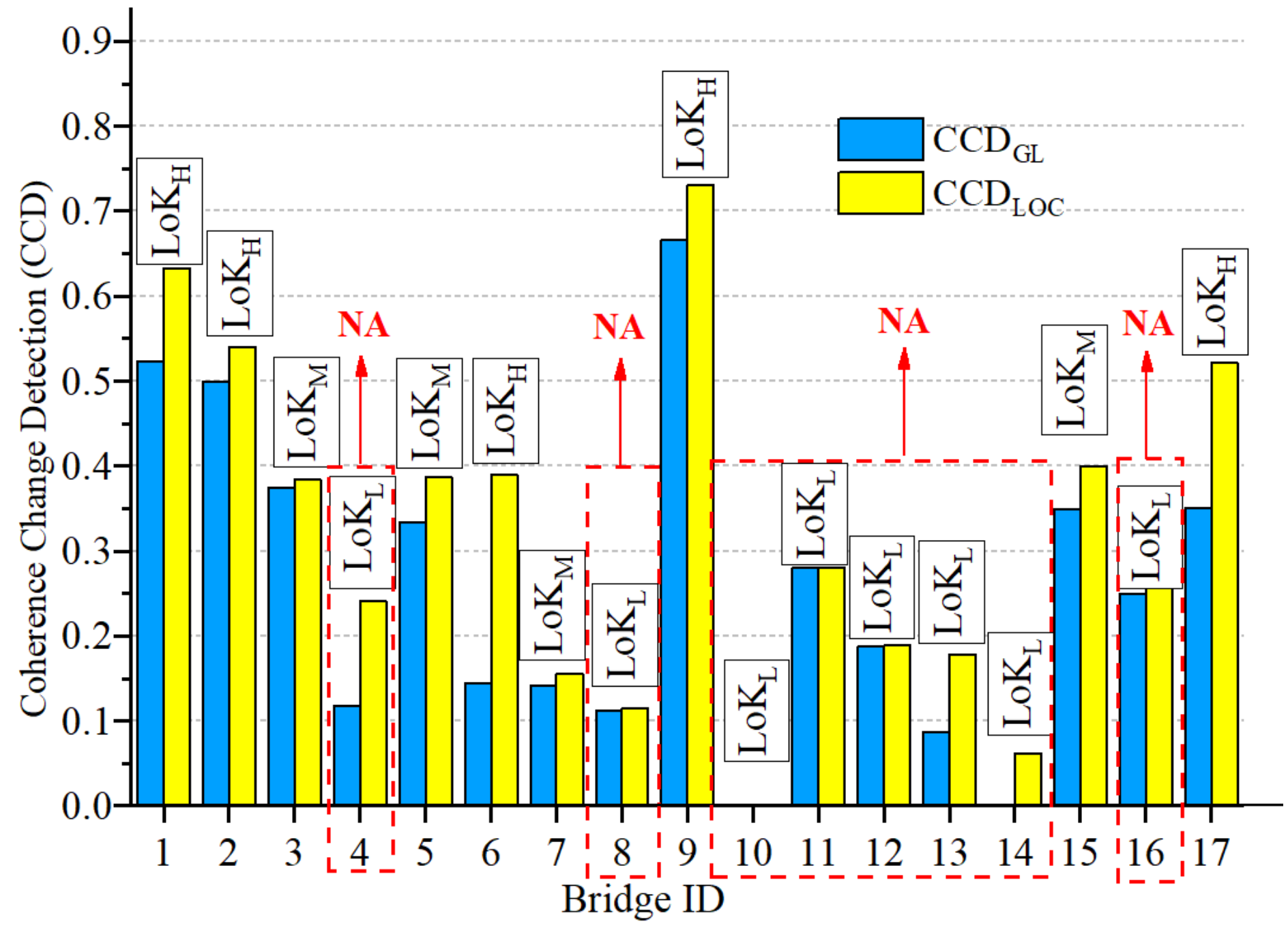

*Figure 9. Coherence Change Detection (CCD) and level of knowledge (LoK) for the bridges studied. $CCD_{LOC}$ values correspond to the maximum CCD for the area of the asset and local damage. $CCD_{GL}$ values correspond to the mean destruction within the area of the analysed structure. NA (not applicable) indicates bridges, for which the estimated coherence between SAR images from the time period 1 (TP1) was low. These assets were classified as those having Low Level of Knowledge ($LoK_L$) and, thus, were excluded from further analysis.*

Based on the CCD values shown in Table 2, the damage of each asset was evaluated. For this purpose, three damage levels (DL) were defined, i.e. $DL_L$ (low), $DL_M$ (moderate), and $DL_H$ (high) based on the $CCD_{LOC}$ and $CCD_{GL}$. Thus, each asset was assigned with an index, linked to its damage level (DL) and bridges can be grouped according to the level of damage (DL) with the use of assumed approximate ranges (see Table 3).

**Table 3.** Damage characterisation of infrastructure assets (bridges) based on CCD.

| DL | $CCD_{LOC}$ (Max) | $CCD_{GL}$ | Colour legend | Description |
|---|---|---|---|---|
| $DL_H$ (High) | 0.5-1.0 | 0.4-1.0 | | **Severe/Complete damage**: Complete destruction of the structure or of some of its components. |
| $DL_M$ (Moderate) | 0.35-0.5 | 0.3-0.4 | | **Moderate/Extensive damage**: Considerable damage in some of the components. |
| $DL_L$ (Low) | 0.0-0.35 | 0.0-0.3 | | **No/Minor damage**: General deterioration, signs of slight damage. |

The maximum coherence changes ($CCD_{LOC}$) signify the greatest changes within the bridge deck plan view area, referring to a small portion of the deck (below 30%). This explains why $CCD_{LOC}$ was coined here to represent local damage, e.g., damage of one span of a multi-span bridge, or damage of one structural component, while rest of the bridge is unaffected (see Figures 10 a,b,c). Local damage is different from global damage that is expressed by $CCD_{GL}$ values (Figures 10 d,e,f). Global damage is the mean coherence difference, that refers to entire plan view of the bridge, i.e. the area visible from the satellite. For instance, low or medium $CCD_{GL}$ results indicate general deterioration of the bridge: road pavement damage, concrete crushing, and spalling. Therefore, different levels at both local and global scales could serve as the prerequisite for evaluation of remaining capacity of the bridge, both structural and traffic.

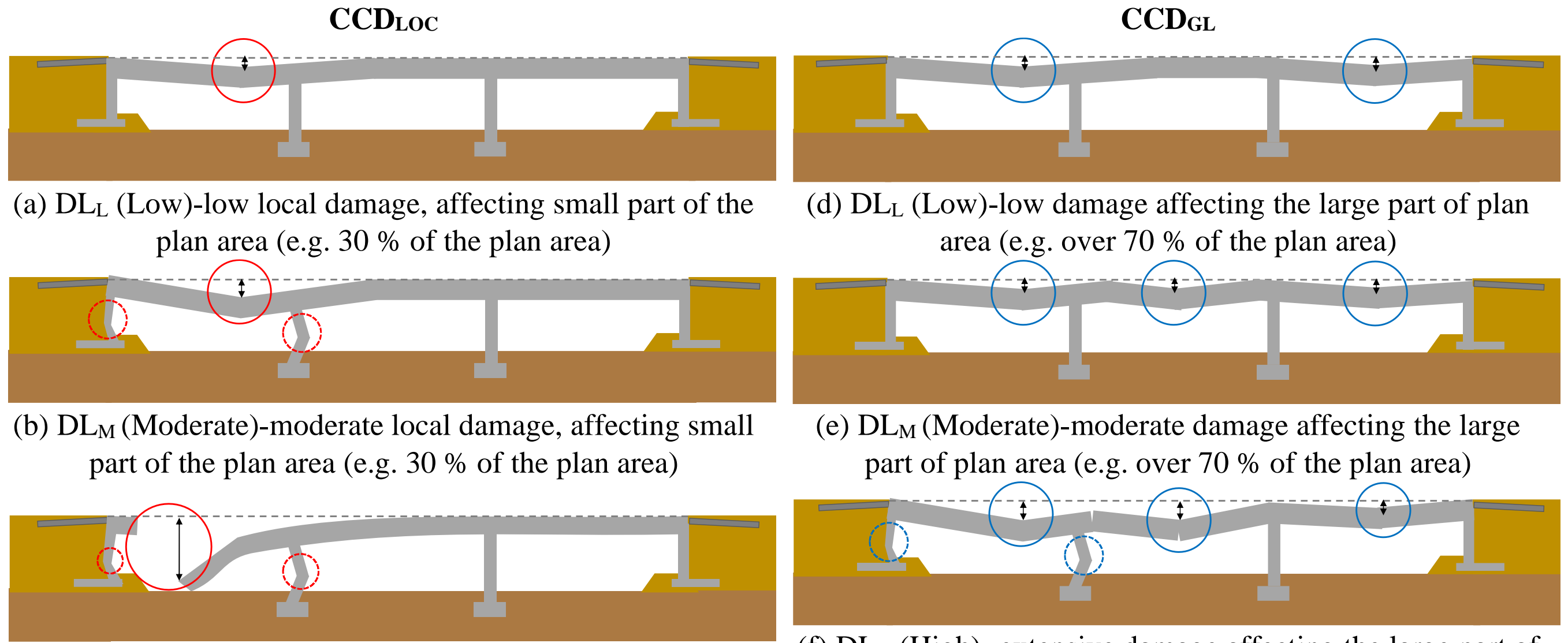

(a) $DL_L$ (Low)-low local damage, affecting small part of the plan area (e.g. 30 % of the plan area)

(d) $DL_L$ (Low)-low damage affecting the large part of plan area (e.g. over 70 % of the plan area)

(b) $DL_M$ (Moderate)-moderate local damage, affecting small part of the plan area (e.g. 30 % of the plan area)

(e) $DL_M$ (Moderate)-moderate damage affecting the large part of plan area (e.g. over 70 % of the plan area)

(c) $DL_H$ (High)-extensive local damage, affecting the small part of the plan area (e.g. 30 % of the plan area)

(f) $DL_H$ (High)- extensive damage affecting the large part of the plan area; global collapse (e.g. over 70 % of the plan area)

*Figure 10. Illustration of different types of damage level (DL), identified by (a,b,c) $CCD_{LOC}$ and (d,e,f) $CCD_{GL}$. Continuous line red circles indicate $CCD_{LOC}$ values, corresponding to local damage and continuous line blue circles outline $CCD_{GL}$ values, associated with global damage that affects most of the area of the bridge deck. The dashed line circles indicate potential damage which may not be verifiable by satellite imagery.*

Changes in coherence within two pre-damage and two pre- and post-damage images were employed for characterising the level of asset damage, as per section 3.2. The results for the $CCD_{LOC}$ and $CCD_{GL}$ between two pairs of images are given in columns 6 and 7 of Table 2. Figure 11 shows the CCD for the analysed bridges. To translate these CCD values into damage, $CCD_{LOC}$ values correspond to the greatest change of coherence, localised in specific areas across the entire plan view area of the asset. Thus, $CCD_{LOC}$ indicates extensive local damage of the bridge. In contrast, high $CCD_{GL}$ values indicate global damage affecting the largest part of the asset plan view area. For example, if a bridge is damaged locally by shelling, which however has not affected the entire bridge, this will lead to a high value of $CCD_{LOC}$ and a lower value of $CCD_{GL}$. To translate these CCD values into damage, $CCD_{LOC}$ values correspond to the greatest change of coherence, localised in specific areas across the entire plan view area of the asset. Thus, $CCD_{LOC}$ indicates extensive local damage of the bridge. In contrast, high $CCD_{GL}$ values indicate global damage affecting the largest part of the asset's plan view area.

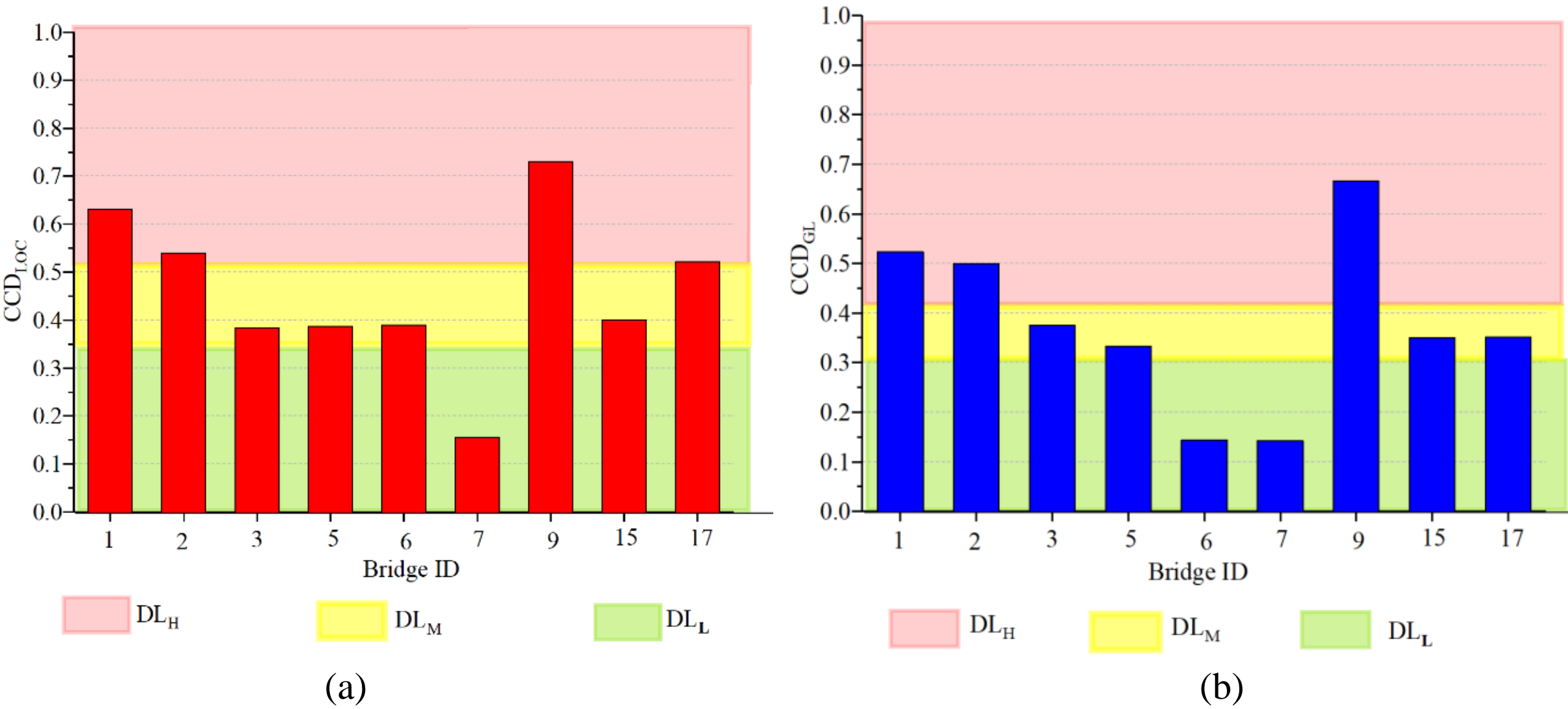

(a) (b)

*Figure 11. CCD values within the estimated period (before and after the disaster): (a) maximum $CCD_{LOC}$ values, corresponding to localised damage of the bridge, (b) $CCD_{GL}$ values, indicating the mean destruction affecting a large part of the area. DL is the damage level characterisation: L (low), M (medium) and H (high) (see Table 3).*

During the analysis, limitations were identified relating to the (a) spatial resolution of the satellite, (b) sequence of events, (c) line of sight. Geospatial datasets and GIS environment were utilised for additional illustration of potentially disrupted areas on bridges identified at asset scale (Appendix A, Figures A.3-A.5). Also, some examples of damage detection and evaluation for assets of $LoK_H$ that have the most extensive destruction ($DL_H$) are shown in Figure 12.

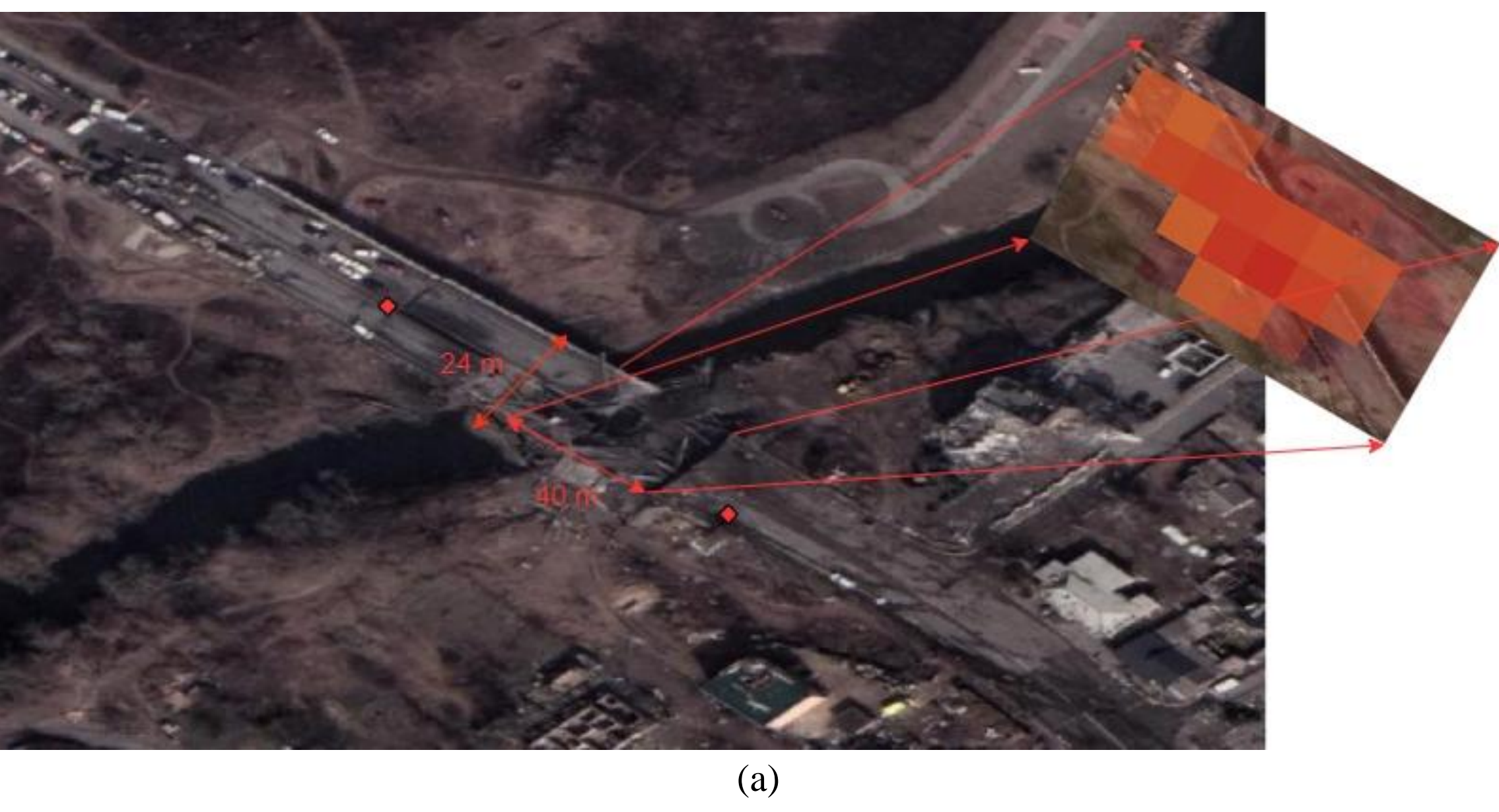

(a)

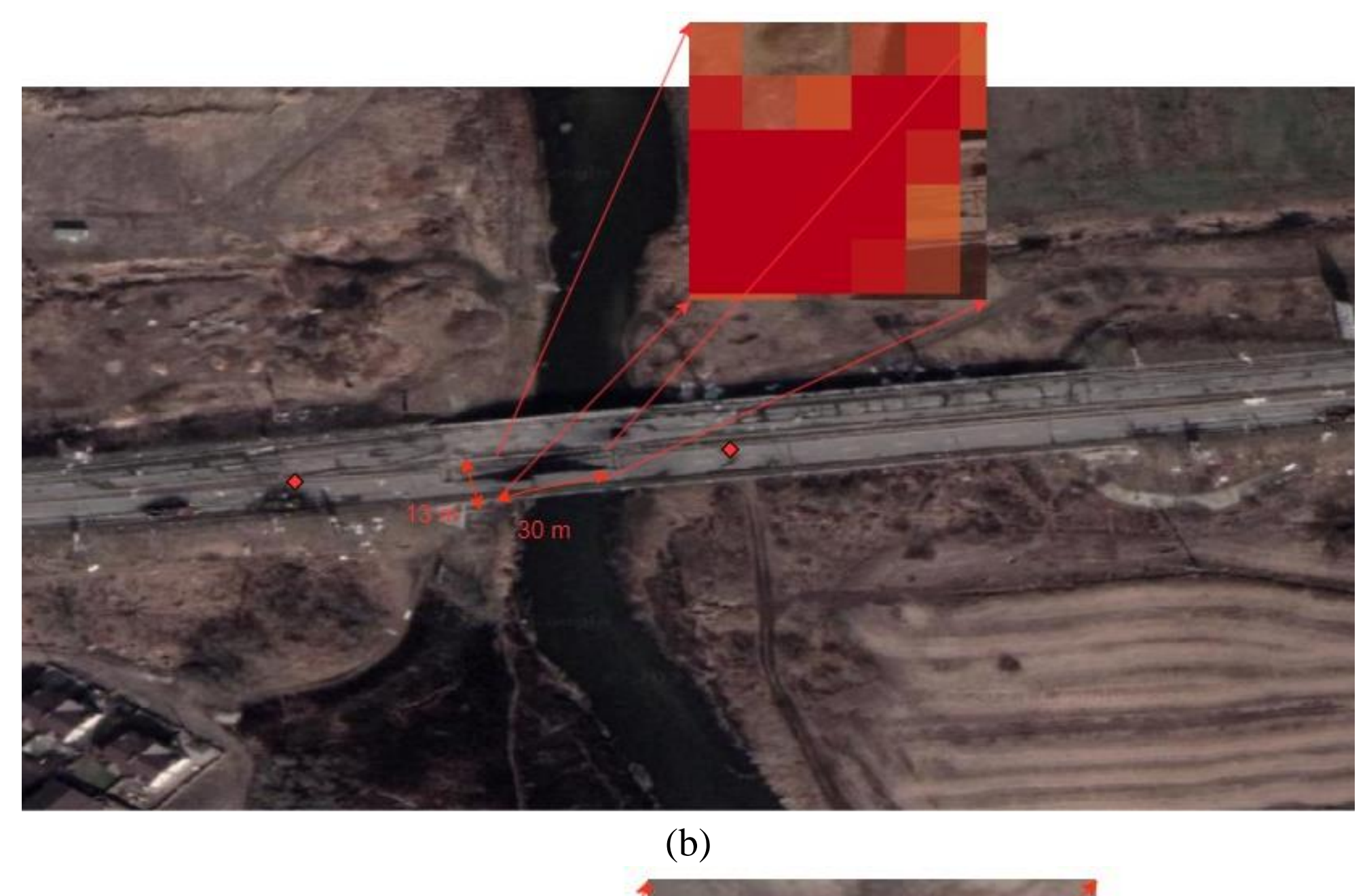

(b)

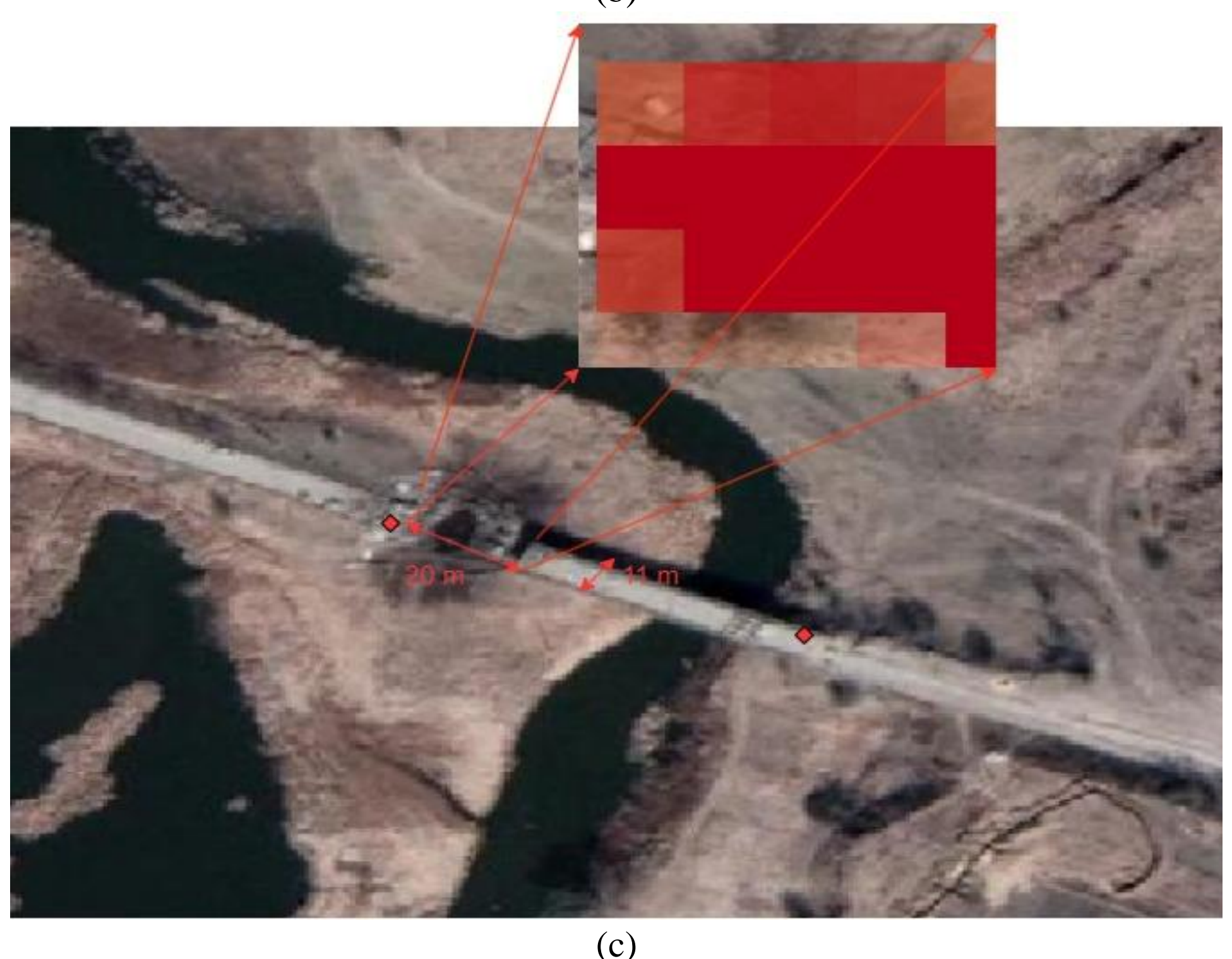

(c)

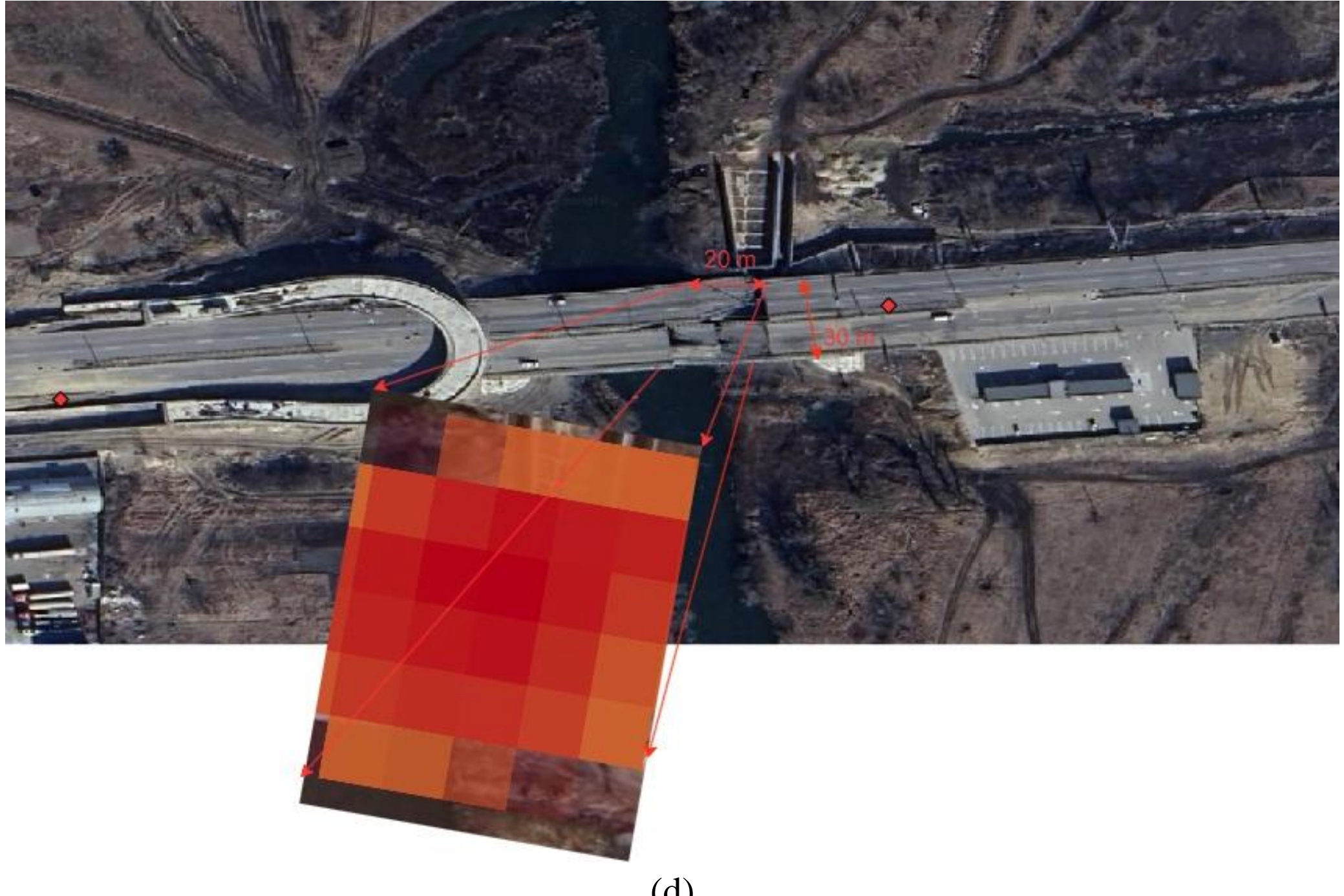

(d)

*Figure 12. Damage characterisation, demonstrating High LoK: (a) B1- $DL_H$, (b) B2- $DL_H$, (c) B9-$DL_H$, (d) B17-$DL_M$. Red points indicate two end nodes of the bridge (see Table A.1 in Appendix A). Approximate dimensions of damaged bridge area are provided. Red pixels highlight zone of the asset with $DL_H$. Additional results for bridges with Low and Medium DL are given in the Appendix A, Figures A.3-A.5*

Geospatial datasets were used for the validation of the damage level at asset scale. The cross-validation applied in this research entails the comparison of the CCD damage characterisation results outlined with the high spatial resolution images obtained by Google Earth Pro. The latter provides details on roads, buildings, and other infrastructure. These data served in this paper as an additional visual validation of the results, as per Figure 13. The images on the left were captured before the damage, while the images on the right were taken from March and April 2022. Thus, visual comparison between photos of assets, obtained before the beginning of shelling (October, February 2022) and after the period of the most extensive destruction in the region (April, March 2022) demonstrate an excellent agreement with the results outlined above.

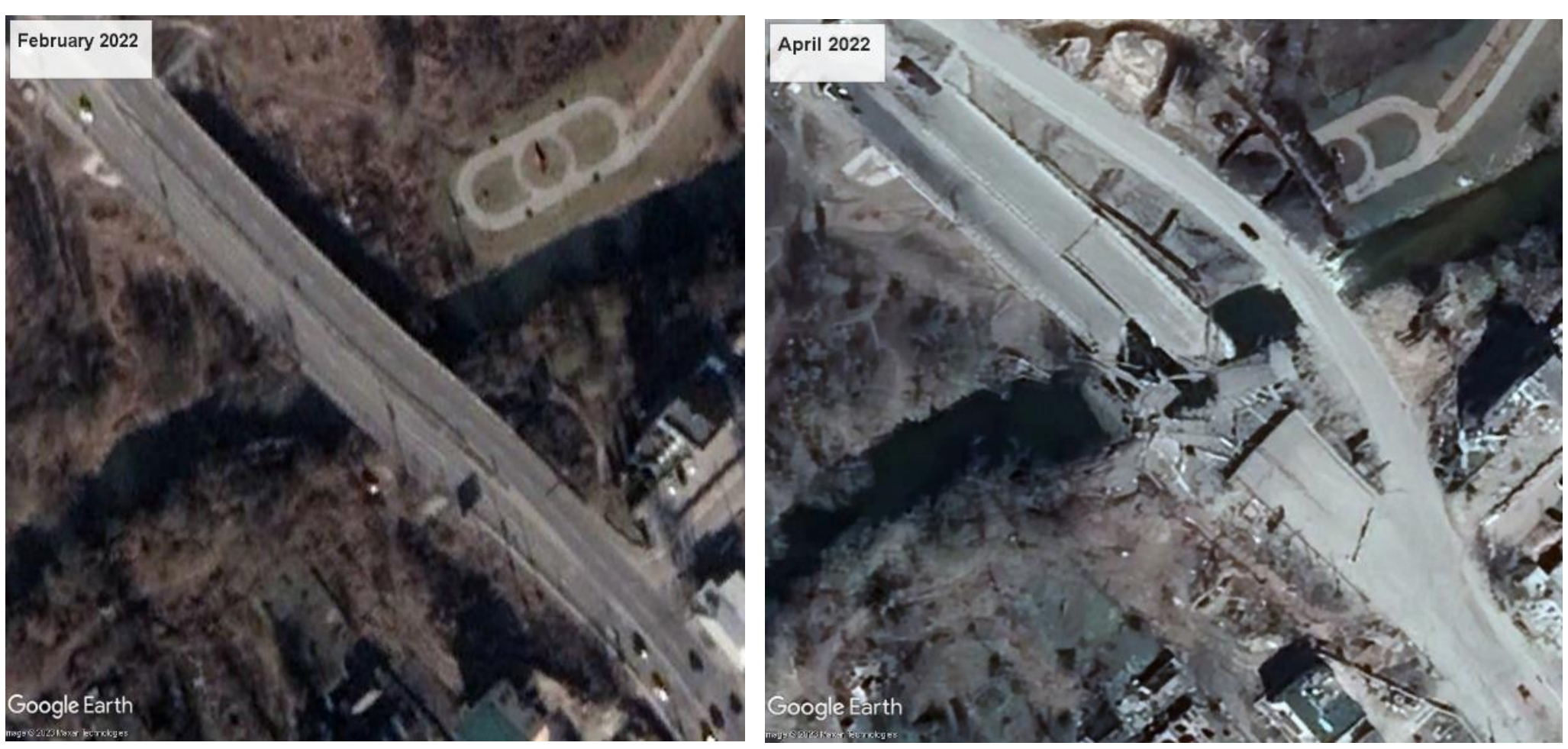

(a) B1: $LoK_H$; $DL_H$; $CCD_{GL}$=0.523; $CCD_{LOC}$=0.623
(on the right a temporary diversion route was constructed to bypass the bridge)

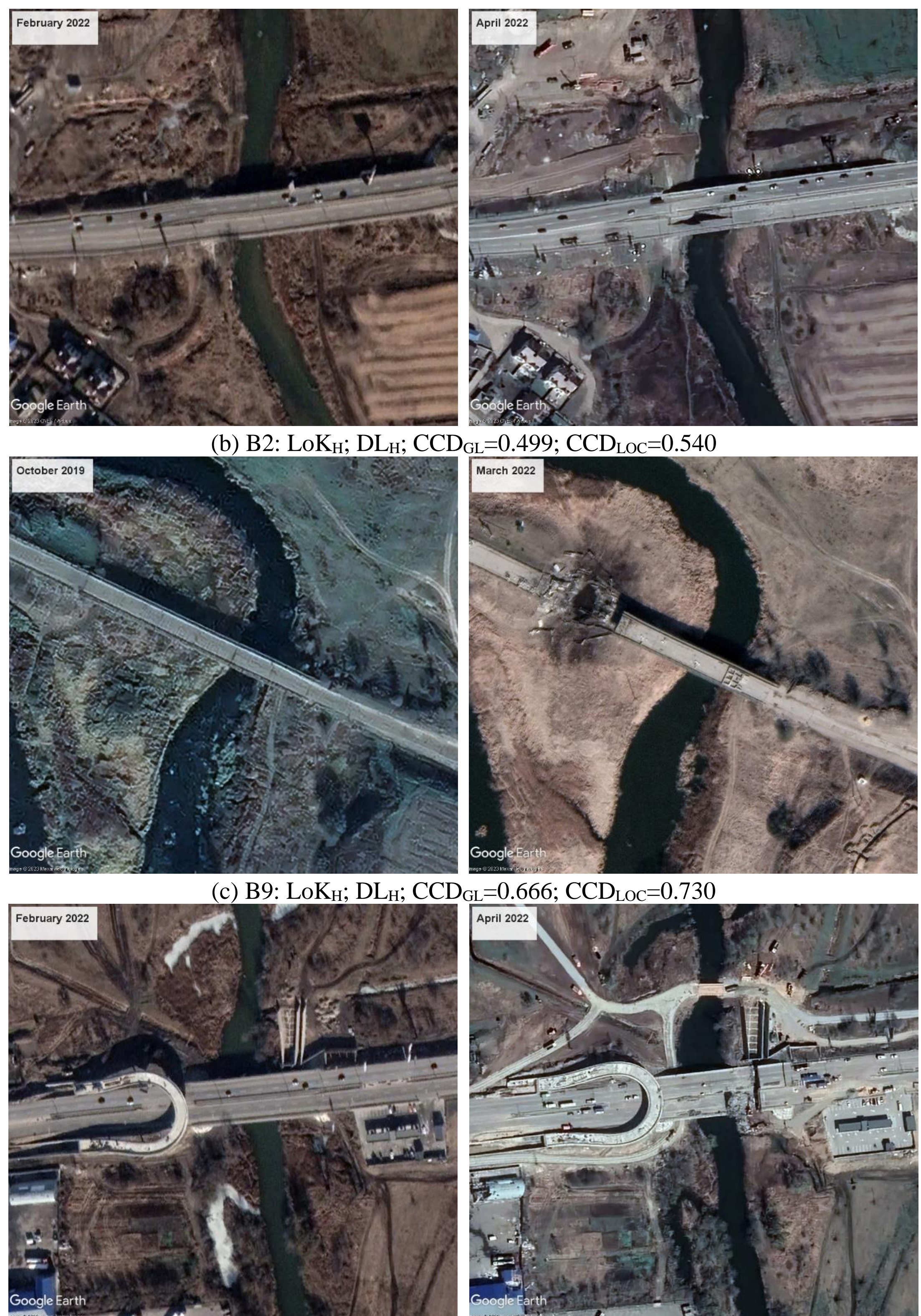

(b) B2: $LoK_H$; $DL_H$; $CCD_{GL}=0.499$; $CCD_{LOC}=0.540$

(c) B9: $LoK_H$; $DL_H$; $CCD_{GL}=0.666$; $CCD_{LOC}=0.730$

(d) B17: $LoK_H$; $DL_M$; $CCD_{GL}=0.351$; $CCD_{LOC}=0.521$ (on the right a temporary diversion route was constructed to bypass the bridge)

*Figure 13. High spatial resolution images from Google Earth Pro used to validate the damage characterisation of bridges B1, B2, B9 and B17 as per Sentinel-1 coherence and CCD products.*

Thus, damage evaluation at asset scale for assets of High LoK enabled to identify zones within the assets, with extensive damage. For instance, by incorporating additional data sources (e.g., from crowdsourcing, geospatial datasets) dimensions of bridges can be estimated (see Table A.1 in Appendix A) and the area (size) of the damaged zone for each of assets can be calculated (see Figure 12). This information is used to facilitate restoration strategies, utilising Goole Earth pro photos at different periods, for the damaged bridges (e.g., Figure 13), depending on, e.g., the location and type of bridge. In addition, Figures 12-13 illustrate useful information regarding the traffic disruption and corresponding impact on infrastructure operability in the region can be obtained from open-access sources. Hence, the extent of damage to a bridge directly correlates with the disruption to traffic flow. For instance, the destruction of B1 (see Figure 12a and Figure 13a) extends across the entire width of the bridge, resulting in the complete disruption of traffic on P30 highway, a regionally significant route traversing the territory of the Kyiv region with a total length of 6.4 km. In particular, the disruption of this transport route resulted in the isolation of a portion of the region from the capital city, Kyiv, leading to significant social and economic repercussions. Damage characterisation of the B1 bridge is investigated in more detail at component level (see section 4.3). Similarly, as B2 and B17 also are vital for the operability of international transport routes, their destruction causes crucial consequences for the logistics. Thus, the open access geospatial datasets were used to analyse the impact of inoperability of each of bridges on overall infrastructure (see more details in Appendix A). It can be summarized that the integration of damage evaluation results from diverse data sources supports efficient restoration planning and execution for hazard-affected regions, aiding in minimizing downtime, optimizing traffic flow, and expediting the reinstatement of traffic capacity through effective stakeholder coordination and implementation of traffic management strategies.

### 4.3 Component scale damage assessment

When asset scale assessment based on CCD is not adequate for designing restoration strategies, damage characterisation at component scale is required. Here, the component scale automatic damage assessment for B1 with $LoK_H$ and $DL_H$ was conducted. In doing so, open access platforms such as Damage In UA [84], the data available at [85] and [86] were used to obtain links to trustworthy sources with images of bridge for damage detection. Using the methodology described in Section 3.4, for the selected bridge B1, two automated computer vision tasks were performed. (a) Instance segmentation of affected bridges for component detection and classification, (b) Instance segmentation for defect detection, location, and classification for the following categories (e.g., crack) damage characterisation.

The outputs of SAM and Grounded-SAM are shown in Figure 14. Firstly, the input images undergo mask extraction using SAM everything mode. The visual representation of SAM segmentation results is showcased in the second row, while the individual mask outcomes are presented in the third row. Subsequently, employing Grounding DINO [93], labels were matched with the obtained masks, excluding masks below the recognition threshold from display, and the results are displayed in the Grounded-SAM row. Additionally, all labels extracted from the masks, and some other descriptive words from Grounding DINO are presented in the ‘tags’ row.

In contrast to the common bridge component recognition research, severe damage conditions introduce two challenges: firstly, the background for detection becomes highly intricate. Secondly, there is uncertainty of occlusion affecting the bridge (see Figure 15a). Therefore, Grounded-SAM is initially employed for an initial general detection task (Figure 15b), with the objective of identifying the specific mask corresponding to the bridge (see Figure 15c). Subsequently, the masks intersecting with the bridge are filtered, pixels within these masks are removed, and Stable Diffusion is employed to fill in the erased areas (Figure 15d). At the same time, the original bridge subject is extracted from the image, and any remaining areas are replaced with a white mask (Figure 15e). This process effectively eliminates intricate backgrounds and occlusions unrelated to the bridge subject. The repaired bridge mask is extracted from the background once again (Figure 15g).

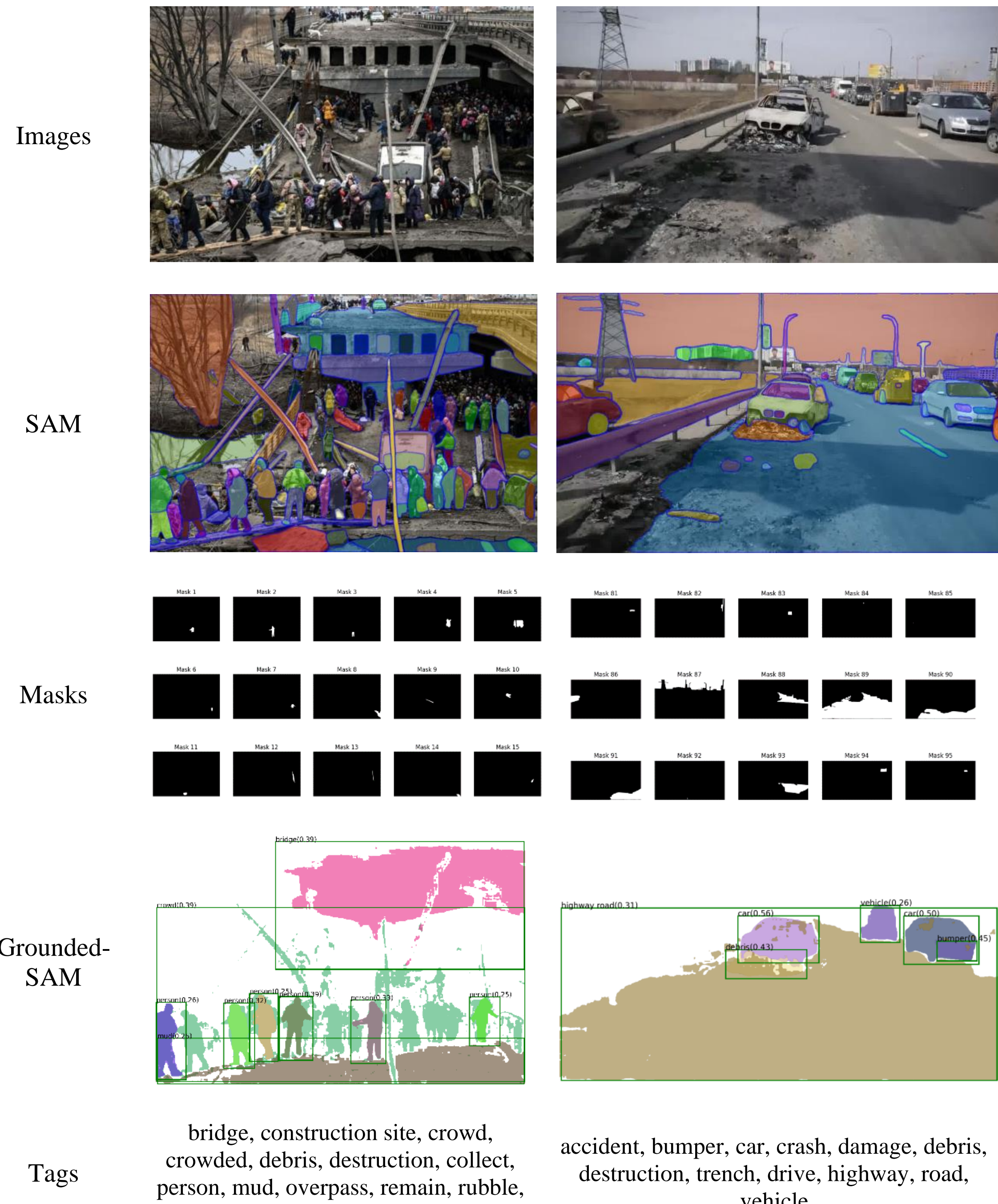

*Figure 14. The outputs of SAM and Grounded-SAM for damage detection (e.g. crack) of bridge at component level.*

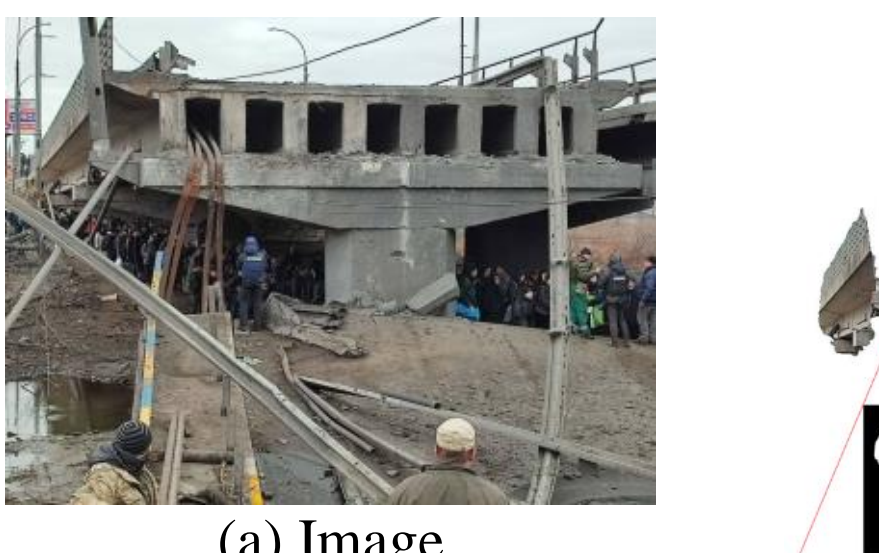

(a) Image

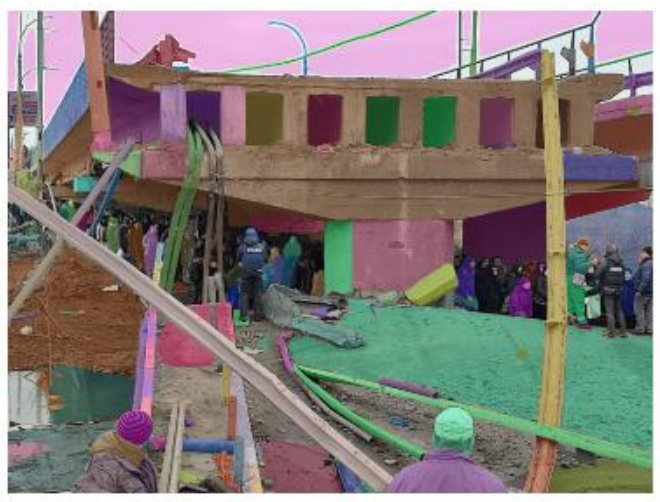

(b) SAM Result

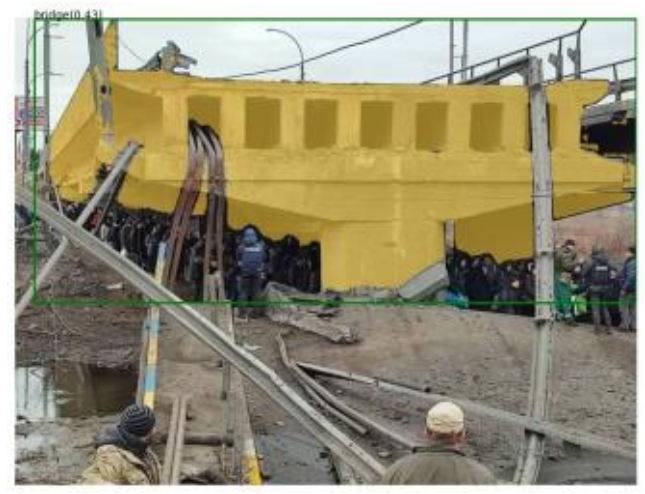

(c) Grounded-SAM (general detection)

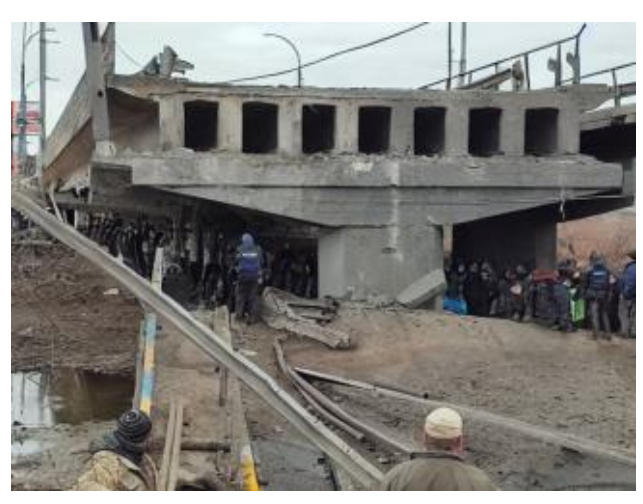

(d) Stable Diffusion for Occlusion Repair

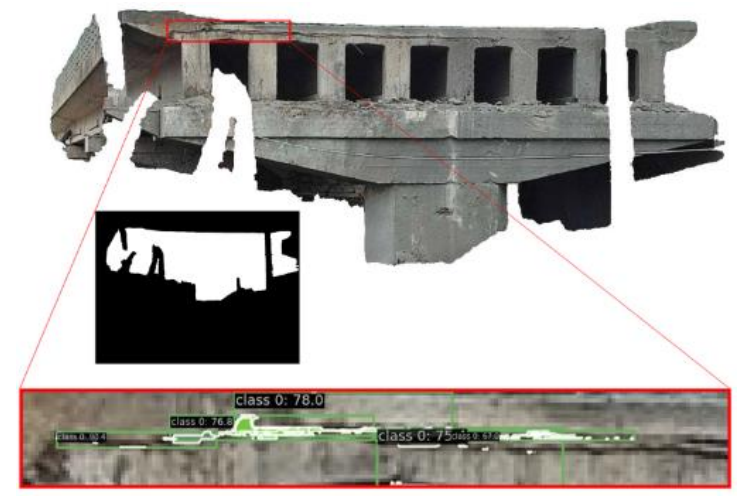
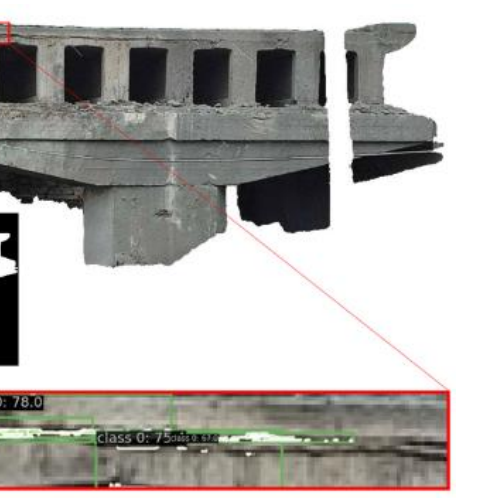

(e) Bridge Capture and Crack Detection

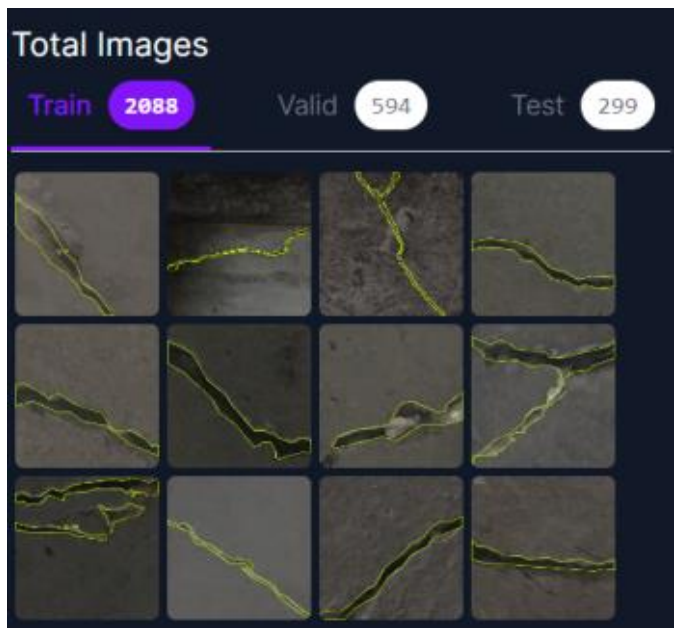

(f) Training Set for Crack Detection [96]

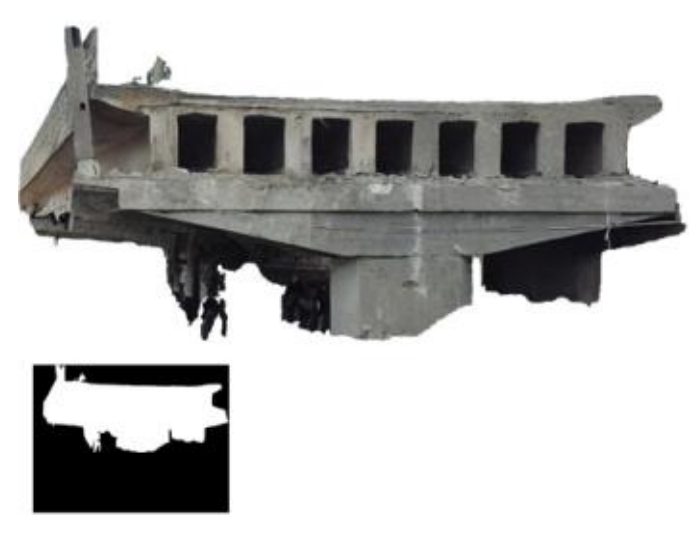
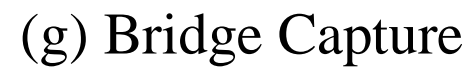

(g) Bridge Capture

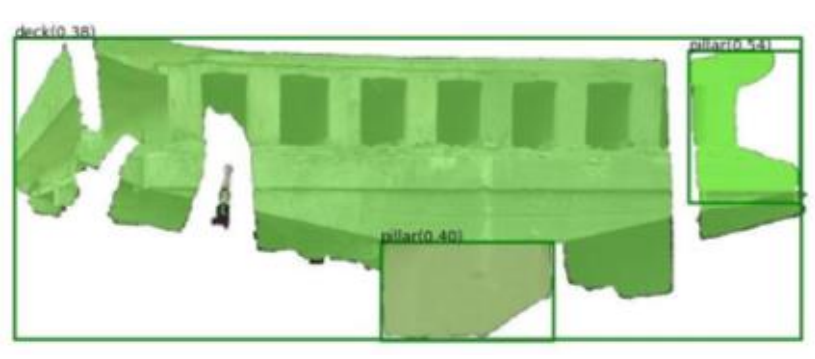

(h) Instance Segmentation for Component Level based on SAM-generated mask

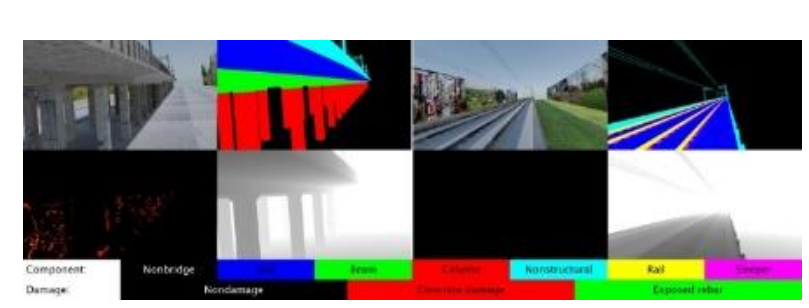

(i) Training Set for Component Level Detection (Tokaido dataset)

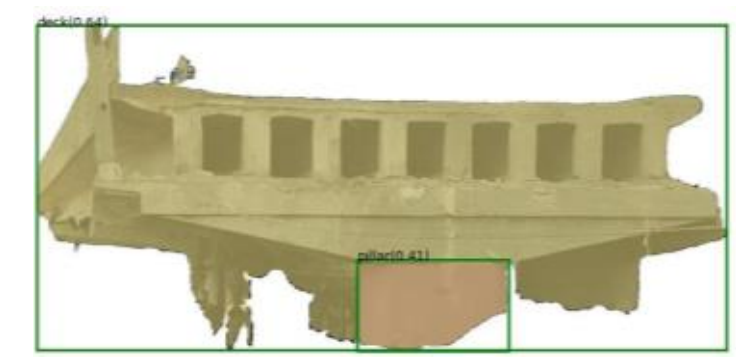

(j) Instance Segmentation for Component Level based on Repaired SAM-generated mask

*Figure 15. Combined component and damage detection for B1 bridge.*

For the extracted bridge masks that has not been repaired, the crack detection task will be performed using a dedicated mode. This crack detection model has been trained on a dataset of approximately 3000 images (Figure 15f) and utilises the query-based learnable prompt SAM algorithm mentioned in section 3.4 [92] as shown in Figure 15e. The occlusion repaired bridge undergoes bridge component detection tasks. The model employed here is also self-prompting SAM, trained on the Tokaido dataset [96], illustrated in Figure 15i. As shown in Figure 15h and 15j, the component detection result significantly improves after the occlusion repairs. Upon completing both the crack detection and bridge component recognition tasks, a simple matching process is carried out to determine the location of the cracks within specific bridge components. This approach enabled automatic damage detection based on the importance of bridge components. For instance, the crack was detected in the critical for structural integrity component (pier) (Figure 15e), providing evidence of the severe deterioration of the asset. Thus, the CV technologies at component scale have shown the outstanding capability to enhance automatic damage detection, following the general tiered multi-scaled approach presented in this paper.

## 5. Conclusions

An integrated multi-tier framework for automatic infrastructure damage characterisation at different scales (regional, asset and component) using openly available data is introduced and applied to damaged infrastructure assets. The proposed methodology facilitates the prompt detection of damage during disaster assessment and rehabilitation efforts. Automatically and precisely detecting and characterising damages following human-induced disasters poses a challenging yet vital endeavour. From regional to asset and to component scales, the assessment of critical infrastructure affected by threats is performed, using for the first time, a novel metric that relies on the change of interferometric coherence (CCD). The new methodology, depending on the level of knowledge, which relates to the reliability of the data, characterises infrastructure damage level based on measurements of CCD, representing either local ($CCD_{LOC}$) or global damage ($CCD_{GL}$). CCD values are then correlated with three distinctive damage levels, to then be validated by high-resolution images. To characterise damage at the component scale, we utilised advanced computer vision techniques. By extracting the mask of the bridge main body and repairing occluded areas, we successfully achieved defect recognition and localisation for each bridge component, enabling comprehensive damage characterisation.

This methodology is valuable for the rapid assessment and decision making for the reconstruction of critical infrastructure. The framework is applicable to regions that have been affected by threats, such as climate hazards and human-induced damage, when the scale of the damage is significant in terms of intensity and extent and/or when accessibility is impossible, e.g., flooded areas or war zones. This is because the only way to restore damaged infrastructure is to know beforehand the size and extend of destruction, and this information can only be provided by damage level characterisation. Limitations in damage assessment at regional and asset scales using Sentinel-1 images were identified including constraints relating to the spatial resolution of the satellite, challenges associated with the sequence of events, consideration regarding the line of sight. However, the feasibility of this approach was substantially increased by integrating disparate data sources for precise timing, elimination of weather and line of site impacts and overcoming considerable class imbalance in urban environments.

Automatic integration of stand-off observations and open-access information from disparate sources into recovery planning was proven to enable an informed response to hazards, facilitating expeditious decision-making processes for infrastructure development and the design of efficient restoration strategies.

### Acknowledgements

The first author would like to acknowledge the financial supports from British Academy for this research (Award Reference: RaR\100770).

Stergios-Aristoteles and Sotirios Argyroudis received funding by the UK Research and Innovation (UKRI) under the UK government's Horizon Europe funding guarantee [Ref: EP/Y003586/1, EP/X037665/1]. This is the funding guarantee for the European Union HORIZON-MSCA-2021-SE-01 [grant agreement No: 101086413] ReCharged - Climate-aware Resilience for Sustainable Critical and interdependent Infrastructure Systems enhanced by emerging Digital Technologies.

Several maps included in this work were created using ArcGIS® software by Esri. ArcGIS® and ArcMap™ are the intellectual property of Esri and are used herein under license. Copyright © Esri. All rights reserved. For more information about Esri® software, please visit https://www.esri.com/en-us/home (accessed on 15 December 2023). The authors are grateful to the European Space Agency and the National Aeronautics and Space Administration, who provided Sentinel-2 and SRTM data accordingly.

### CRediT authorship contribution statement

Nadiia Kopiika: Methodology, Validation, Formal analysis, Investigation, Writing – original draft, review & editing; Andreas Karavias, Pavlos Krassakis: Conceptualisation, Methodology, Writing –review & editing; Visualisation; Nataliya Shakhovska: Conceptualisation; Zehao Ye: Methodology, Formal analysis, Investigation, Writing – original draft, Visualisation; Jelena Ninic: Conceptualisation, Methodology, Writing – original draft, review &

editing, Supervision; Nikolaos Koukouzas: Conceptualisation; Sotirios Argyroudis, Stergios-Aristoteles Mitoulis: Methodology, Conceptualisation, Writing –review & editing, Supervision;

**Declaration of Competing Interest**

The authors declare that they have no known competing financial interests or personal relationships that could have appeared to influence the work reported in this paper.

# Appendix A

## A.1. The research-related nomenclature and open-access data sources

**Human-induced hazards/Anthropogenic disasters** (this paper) are catastrophic events caused or significantly influenced by human activities, causing potential threats to the environment, the society, and infrastructure. This study is focused on the group of such accidents, integrating destructions caused by terrorist attacks, military activities, and hostilities at conflict-prone territories. These hazards can have severe consequences for human populations, infrastructure, and the environment, encompassing a range of threats, and their impacts can be both immediate and long-lasting. Addressing war-induced hazards requires comprehensive efforts, including conflict prevention, peacebuilding, humanitarian assistance, and post-conflict reconstruction.

**The Level of Knowledge (LoK)** (in this paper): the proposed parameter for estimation of reliability of damage detection at asset scale. As the study is focused on sources of data, freely accessible during the hostilities, the utilised Sentinel-1 imaginary reveal certain limitations. Thus, all the assets were classified, according to three Levels of Knowledge: low ($LoK_L$), medium ($LoK_M$), and high ($LoK_H$), identifying the applicability of the approach. Classification of assets was mostly based on the engineering judgement and is the first of such kind in international literature. The main principle of the proposed data quality assessment and classification was laid in assessment of coherence values between 2 images for the first time period (TP), covering close datasets. Although such approach is not widely used, there some studies, aiming to assess the quality of the Sentinel data according to coherence between 2 images [88],[102],[103],[104]

**The Damage Level (DL)** (this paper): the proposed classification of assets based on the change of coherence between the pairs of images from two datasets. Structures were classified in damage levels: $DL_L$ (low), $DL_M$ (moderate), $DL_H$ (high).

**Coherence products** analysed in this case study:

- **local** ($\gamma_{LOC}$)- indicates the maximum coherence, which was possible to indicate for the pair of images;
- **global** ($\gamma_{GL}$)- indicates the range, for which 95% of the data is within two standard deviations ($2\sigma$) of the mean value.

The deployment of two types of products was motivated by the fact that the coherence of an image varies from area to area or even between individual pixels. Hence, the coherence image can be used to assess the quality variation of an interferogram over the analysed area, similarly as discussed in [102]. Thus, the brighter areas on the coherence products correspond to higher coherence (e.g. $\gamma>0.7$) and the dark areas correspond to lower coherence (e.g. $\gamma<0.5$) (see Fig. A.1). To ensure the substantial quality of the products utilised, the histogram of the coherence image is used, demonstrating which ratio of pixels in the assessed area fall under the high (medium) coherence level (e.g. $\gamma_{GL}>0.7(0.5)$), thus being applicable for damage evaluation. At this stage, possible areas with low coherence ($\gamma_{GL}<0.5$) can be identified and neglected from further processing and damage evaluation. (e.g. Figure A.1.b)).

## A.2. Open-access data sources for crowdsourcing

**OpenStreetMap (OSM)** [83] is a publicly accessible dataset providing geospatial data in global scale that are related with land uses, transportation networks, and infrastructures. This dataset undergoes continuous updates from users worldwide, making it an important source of geospatial information for both commercial and research applications. Launched in 2004, OSM allows anyone contribute by adding new data, correcting existing information, or enhancing details about specific locations. OSM plays a crucial role in humanitarian efforts by providing up-to-date maps for disaster-stricken areas. Volunteers often contribute by mapping affected regions to aid disaster response and recovery.

**Damage In UA** us a project that collects, evaluates, and analyses information on material losses of citizens and the state from the war with Russia. Since the first days of the war, in February 2022, the project has been implemented by the Kyiv School of Economics (KSE) in cooperation with the Office of the President of Ukraine, the Ministry of Economy, the Ministry of Reintegration of the Temporarily Occupied Territories, and the Ministry of Infrastructure of Ukraine[84].

**The Eyes on Russia. The Centre for Information Resilience (CIR)** is a nonprofit social enterprise committed to combating disinformation, exposing human rights abuses, and addressing online behaviour harmful to women and minorities. In January 2022, CIR initiated the **Eyes on Russia project** with the purpose of gathering and verifying various media types, including videos, photos, and satellite imagery, related to the war in Ukraine. The primary objective was to offer journalists, NGOs, policymakers, and the public access to authenticated and trustworthy information. Since its inception, the Eyes on Russia project has facilitated collaborative research within the broader OSINT (Open-Source Intelligence) community, including entities such as Bellingcat and GeoConfirmed, with the support of Advance Democracy, Inc. The database housing verified information is a collective effort of this community. The verified information is compiled in a database and presented on the Russia-Ukraine Monitor Map, with the singular aim of delivering timely and reliable information on the repercussions of the war and its people. In an effort to enhance the functionality of the original map, initially developed and maintained with assistance from MapHub, CIR partnered with C4ADS to create an updated version of the Eyes on Russia Map. C4ADS, a nonprofit organization dedicated to countering illicit networks that pose threats to global peace and security, collaborated with the Eyes on Russia project to produce a new iteration of the map. This updated version aims to expand the capabilities for researchers, allowing them to set search terms and interact with the map in ways that advance their analytical efforts. [85]

**UADamage** is an AI-driven Geographic Information System (GIS) platform designed for the automated analysis of remote sensing data obtained from satellites and drones. Employing computer vision techniques, the platform identifies building boundaries within images and assesses the extent of damage by segmenting each point in the drone or satellite imagery. An impressive technological advancement is achieved through the determination of the building height. The platform calculates the relative height of each pixel within the building in the image. Based on the damage category, the combination of pixel height and building area parameters enables the calculation of the volume for each individual structure [86].

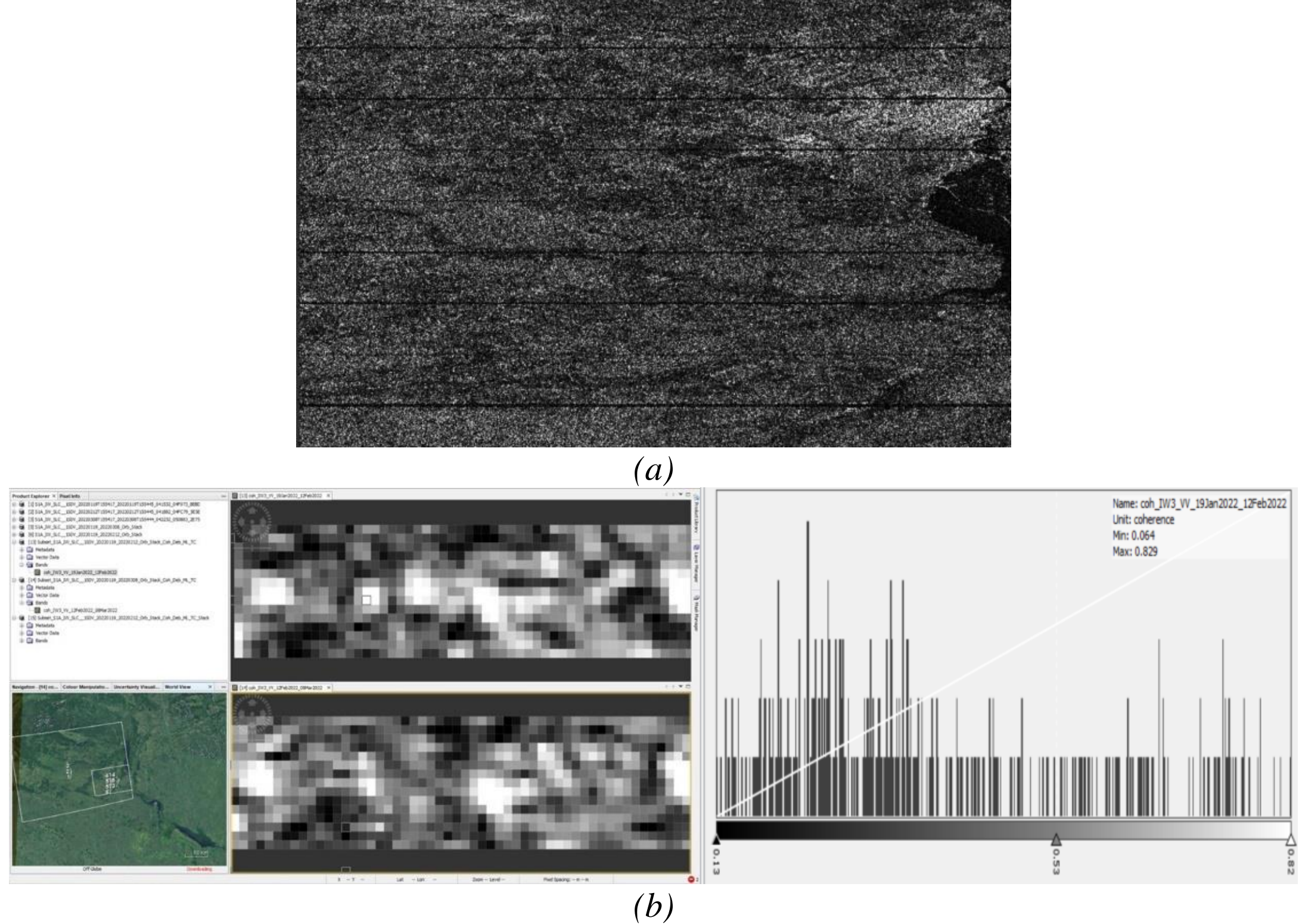

*(a)*

*(b)*

***Figure A.1.*** *Coherence products for evaluation of satellite imaginary quality for damage evaluation: (a) coherence products for the whole area of the Sentinel image and (b) corresponding coherence values and histogram for the localised area of the asset (after subset application). E.g.in (b) most of the pixels within the assessed area have values of lower than 0.5, while the maximum $\gamma_{LOC}$=0.85). Thus, the products do not have the substantial quality for damage evaluation.*

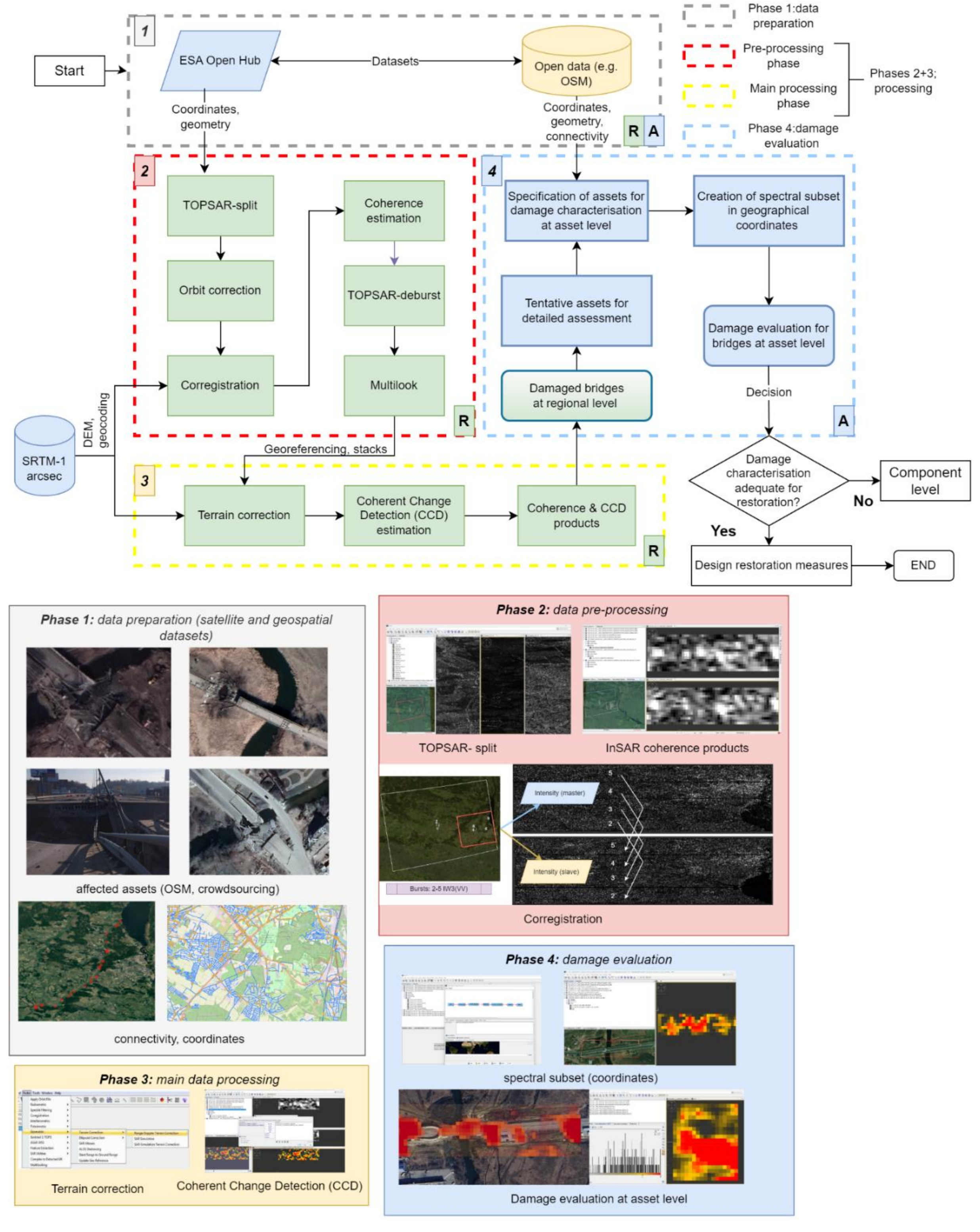

***Figure A.2.*** *Workflow for damage characterisation at regional (R) and asset (A) scale based on four phases:* ***Phase 1*** *(grey-dashed box): data preparation with the use of satellite and geospatial datasets, including location and geometry data of critical assets, residing within the boundaries of the selected study area;* ***Phase 2*** *(red-dashed box): data pre-processing using satellite datasets, including generation of the Coherence products (e.g. Sentinel-1 SAR SLC images);* ***Phase 3*** *(yellow-dashed box): main data processing, including estimation of Coherent Change Detection (CCD) and development of a semi-automated method for the damage detection on infrastructure assets, e.g. bridges; and* ***Phase 4*** *(blue-dashed box): damage evaluation at asset scale using CCD.*

**Table A.1.** List of assets, coordinates, and types of structures in the case study area

| Asset ID | Length/Width (m) | Type | Lon, Lat | |
|---|---|---|---|---|
| | | | end node 1 | end node 2 |
| 1 | 90/24 | Bridge | 50°29'29.680" N<br>30°15'28.934" E | 50°29'27.063" N<br>30°15'33.716" E |
| 2 | 140/27 | Bridge | 50°33'12.613" N<br>30°17'8.608" E | 50°33'12.017" N<br>30°17'2.319" E |
| 3 | 85/10 | Bridge | 50°23'28.229" N<br>30°13'5.070" E | 50°23'28.590" N<br>30°13'3.212" E |
| 4 | 35/8 | Bridge | 50°39'51.805" N<br>30°16'51.514" E | 50°39'52.292" N<br>30°16'50.869" E |
| 5 | 36/9.9 | Bridge | 50°11'50.698" N<br>29°50'10.434" E | 50°11'52.097" N<br>29°50'10.523" E |
| 6 | 155/10 | Bridge +Dam | 50°44'36.703" N<br>30°22'8.149" E | 50°44'40.140" N<br>"30°22'6.879" E |
| 7 | 41/9 | Bridge | 50°36'39.687" N<br>30°16'50.213" E | 50°36'39.872" N<br>30°16'49.264" E |
| 8 | 60/19 | Bridge | 50°42'44.851" N<br>30°20'22.571" E | 50°42'49.559" N<br>30B°20'19.429"E |
| 9 | 87/11 | Bridge | 50°15'0.959" N<br>29°59'59.243 " E | 50°15'1.344" N<br>29°59'58.128" E |
| 10 | 34/4.5 | Bridge + Embankment | 50°18'11.437" N<br>30°4'49.621" E | 50°18'12.102" N<br>30°4'49.269" E |
| 11 | 25/4.2 | Bridge + Weir | 50°27'25.228" N<br>30°14'12.463" E | 50°27'24.733" N<br>30°14'13.572" E |
| 12 | 23/7 | Bridge + Embankment | 50°16'20.038" N<br>30°2'32.858" E | 50°16'20.753" N<br>30°2'32.248" E |
| 13 | 24/2 | Bridge | 50°22'49.958" N<br>30°11'12.009" E | 50°22'50.667" N<br>30°11'12.114" E |
| 14 | 25/3 | Bridge | 50°17'16.262" N<br>30°3'31.742" E | 50°17'16.219" N<br>30°3'33.732" E |
| 15 | 22/8 | Bridge | 50°12'52.507" N<br>29°52'49.863" E | 50°12'52.984" N<br>29°52'49.809" E |
| 16 | 15/4 | Bridge | 50°12'26.815" N<br>29°57'31.024" E | 50°12'27.351" N<br>29°57'31.782" E |
| 17 | 173/30 | Bridge | 50°26'50.775" N<br>30°14'7.284" E | 50°26'50.593" N<br>30°14'4.834" E |
| 18 | - | - | 50°11'50.983" N<br>29°50'13.511" E | 50°11'50.119" N<br>29°50'13.293" E |
| 19 | - | - | 50°20'4.443" N<br>30°8'49.324" E | 50°20'4.432" N<br>30°8'48.161" E |
| 20 | - | - | 50°23'49.212" N<br>30°13'0.984"E | 50°23'49.561" N<br>30°13'1.949" E |
| 21 | - | - | 50°33'44.500" N<br>30°17'1.994"E | 50°33'44.521" N<br>30°17'3.120" E |
| 22 | 9/1.5 | Culvert | 50°12'47.631" N<br>29°50'16.278" E | 50°12'48.601" N<br>29°50'15.890" E |
| 23 | 5/3 | Culvert | 50°12'41.549" N<br>29°50'11.457" E | 50°12'41.876" N<br>29°50'10.381" E |
| 24 | 7/3 | Culvert | 50°12'53.803" N<br>29°52'6.471" E | 50°12'53.443" N<br>29°52'6.631" E |

## A.3. Illustration of different LoK and DL

This section includes details on damage assessment of 17 bridges (asset scale) at different levels of data reliability (LoK) and DL. Figure A.3 shows high level of reliability of results (assets with High LoK), Figure A.4 demonstrates medium level of reliability of results (assets with Medium LoK). Although assets with the lowest LoK (due to low resolution or small size) were excluded from analysis in the paper, they are presented in Figure A.5 in order to illustrate the possible limitation of the method.

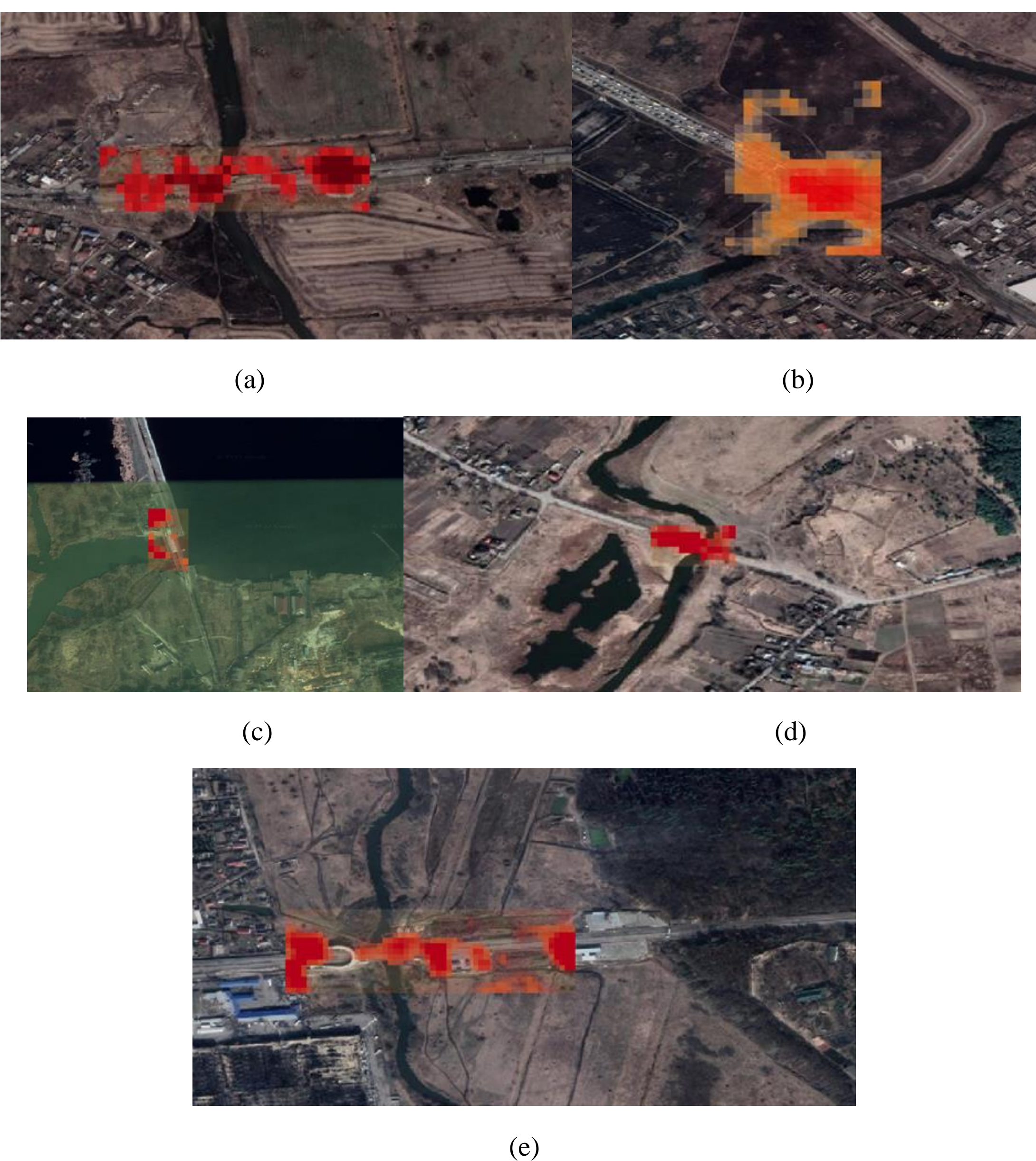

(a) (b)

(c) (d)

(e)

***Figure A.3***. *Damage evaluation at asset scale with $LoK_H$: (a) B1 with $DL_H$; (b) B2 with $DL_H$; (c) B6 with $DL_L$; (d) B9 with $DL_H$; (e) B17 with $DL_H$*

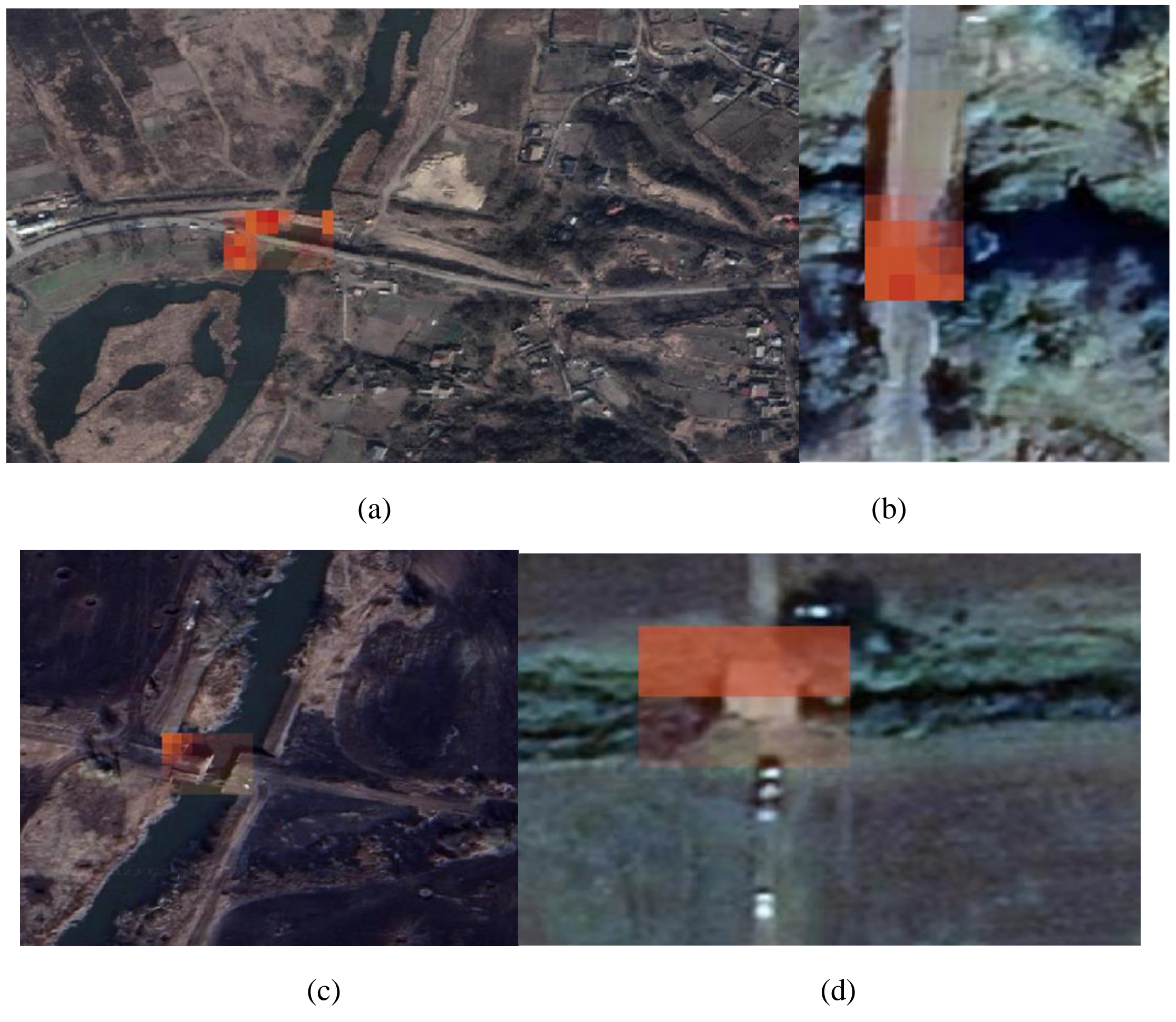

***Figure A.4.*** *Damage evaluation at asset scale with $LoK_M$: (a) B3 with $DL_M$; (b) B5 with $DL_L$; (c) B7 with $DL_L$; (d) B15 with $DL_M$*

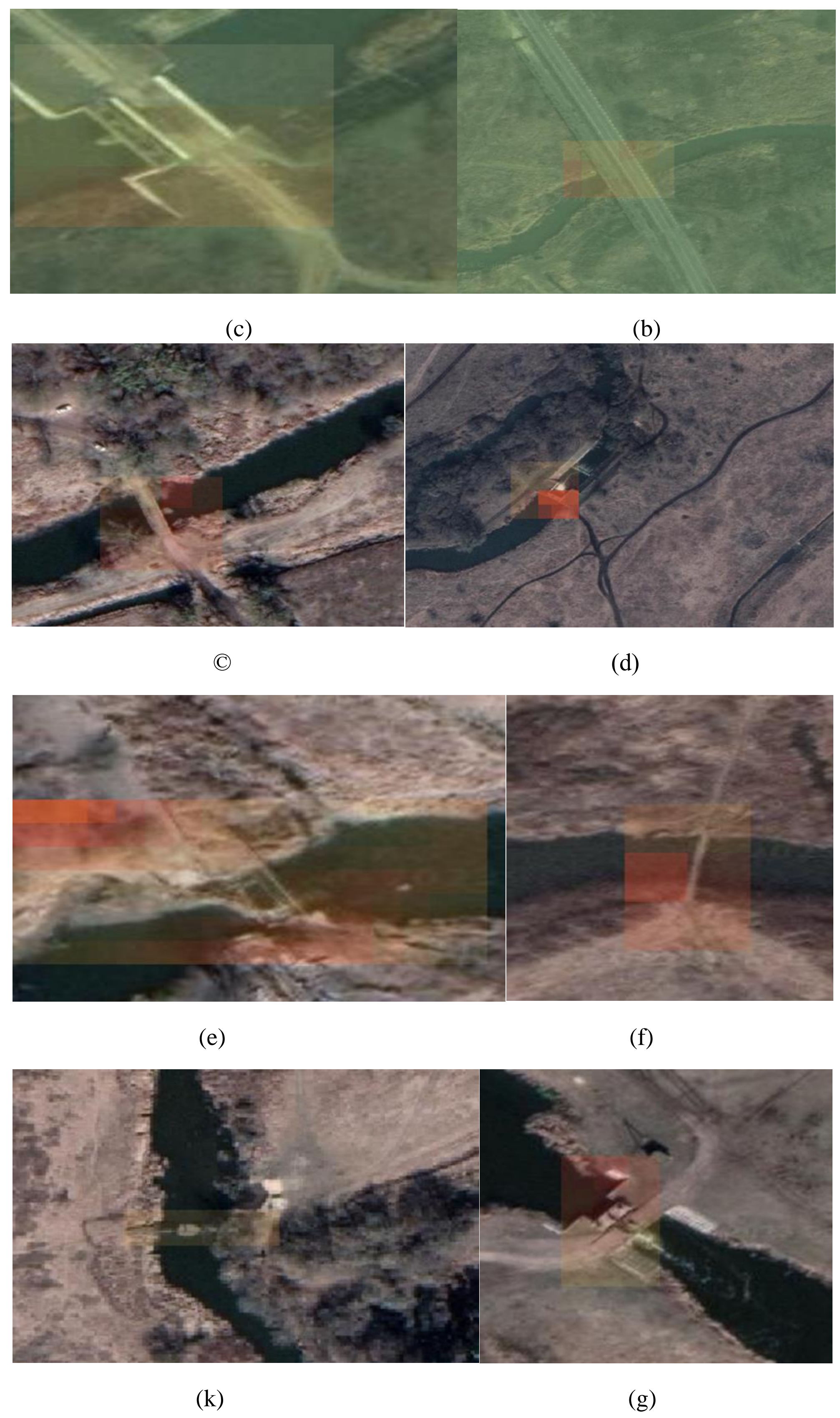

***Figure A.5.*** *Cases with $LoK_L$ at asset scale, which demonstrate the limitation in applying the methodology for damage assessment: (a) B4, (b) B8; (c) B10; (d) B11; (e) B12; (f) B13; (k) B14; (g) B16.*

### A.4 Analysis of assets damage impact on infrastructure operability (example for B1, B2, B9, B17)

Integration of information from disparate open access data sources provides reliable evidence-based prioritisation strategies and decision making for the restoration of entire regions. Here a more detailed discussion is given on the analysis, which can be performed based on Figures 12 and 13 (see paper).

For instance, the destruction of B1 (see Figure 12a and Figure 13a extends across the entire width of the bridge, resulting in the complete disruption of traffic on P30 highway, a regionally significant route traversing the territory of the Kyiv region with a total length of 6.4 km. In particular, the violation of this transport route resulted in the isolation of a portion of the region from the capital city, Kyiv, leading to significant social and economic repercussions. Damage characterisation of the B1 bridge is investigated in more detail at component level (see section 4.3).

Bridge B2 ensures the operability of: (i) M07 highway of international importance, 496.7 km long, connecting Kyiv, Kovel and checkpoint "Yagodin" (border with Poland) and (ii) the European road route E373, passing through the territory of Ukraine and Poland, connecting Kyiv, Korosten, Sarny, Kovel, Yagodyn (Ukraine), Dorogusk, Kholm, Piaski, Lublin (Poland). Thus, the destruction of this asset can lead to the capital of Ukraine being cut-off from an international transport corridor of extreme importance and severely disrupt the busy logistic route, which is internationally important. However, the damage assessment of B2 at asset scale (see Figure 12b and Figure 13b), suggests that only one traffic lane is affected. Therefore, this route can still be used to a limited extent to fulfil logistical requirements for economic and social sectors.

For B9 (see Figure 12c and Figure 13c), it is evident that although the damage level for this asset is high, the destruction is actually located near the bridge abutment, which is in the coastal zone. Hence, the restoration process for this asset is expected to be relatively easy. Also, it is noteworthy, that no traffic routes of regional or international importance pass through this bridge; instead, it mainly serves to connect small towns in the Kyiv region such as Yablunivka, Pereviz, and Leonivka. Given these factors, it can be assumed that the restoration of this asset is of lower priority in the overall rehabilitation strategy of the region, as the closure of this route is likely to result in lower indirect losses.

Finally, bridge B17 serves as a critical passage for the longest European highway, E40, 8,500 km long, connecting the French city of Calais through Belgium, Germany, Poland, Ukraine, Russia, Kazakhstan, Uzbekistan, Turkmenistan and Kyrgyzstan with the Kazakh city of Ridder. It is obvious that the disruption to operability of this asset can have significant impacts on trade, tourism, and overall economic activity, underscoring its critical importance. Hence, from the damage evaluation at asset and regional scale the damage on the bridge deck was found on both lanes, thus causing the complete closure of the route. The deteriorated zone covers a comparatively small portion of the bridge area ($DL_M$-see Figure 12d and Figure 13d); thus, it is likely that the restoration costs and downtime will be lower in this case and the emergency restoration measures can significantly reduce indirect loses.

The application of damage evaluation results for decision making and prioritisation when developing the restoration strategy for the entire region, affected by the hazard, provided in this section, can be potentially utilised for other assets (e.g., buildings, structures). Such preliminary damage detection and evaluation at regional and asset level by combining disparate open-access data sources significantly facilitates rehabilitation process, minimising downtime and eliminating the impact on traffic flow and economic activities. Efficient planning and execution of restoration work, along with effective coordination between stakeholders, help expedite the reinstatement of traffic capacity. Additionally, measures may be taken to optimise traffic flow during the restoration period. This could include implementing alternative routes, temporary bypasses, or traffic management strategies to mitigate congestion and delays.

## References

[1] Sacks R, Kedar A, Borrmann A, Ma L, Brilakis I, Hüthwohl P, Daum S, Kattel U, Yosef R, Liebich T, Burcu Barutcu BE, Muhic S. SeeBridge as next generation bridge inspection: Overview, information delivery manual and model view definition. Automation in Construction, 2018; 90: 134-145. https://doi.org/10.1016/j.autcon.2018.02.033

[2] Argyroudis S A, Nasiopoulos G, Mantadakis N, Mitoulis S A. Cost-based resilience assessment of bridges subjected to earthquakes. International Journal of Disaster Resilience in the Built, Environment, 2020; 12(2). https://doi.org/10.1108/IJDRBE-02-2020-0014

[3] Mitoulis S A, Domaneschi M, Cimellaro G-P, Casas J-R. Bridge and transport network resilience – a perspective, ICE – Bridge Engineering -Themed Issue on Bridge and transport network resilience, 2021; 175(3), 138-149. https://doi.org/10.1680/jbren.21.00055

[4] Macchiarulo V, Giardina G, Milillo P, Aktas Y D, Whitworth M R Z. Integrating post-event very high resolution SAR imagery and machine learning for building-level earthquake damage assessment. Bull Earthquake Eng, 2024. https://doi.org/10.1007/s10518-024-01877-1

[5] Achillopoulou D V, Mitoulis S A, Argyroudis S A, Wang Y. Monitoring of transport infrastructure exposed to multiple hazards: A roadmap for building resilience. Science of the Total Environment, 2020; 746, 141001. https://doi.org/10.1016/j.scitotenv.2020.141001

[6] Mitoulis SA, Argyroudis SA, Panteli M, Fuggini C, Valkaniotis S, Hynes W, Linkov I. Conflict resilience framework for critical infrastructure peacebuilding. Sustainable Cities and Society, 2023; 91, 104405. https://doi.org/10.1016/j.scs.2023.104405

[7] Macchiarulo V, Milillo P, Blenkinsopp C, Giardina G. Monitoring deformations of infrastructure networks: A fully automated GIS integration and analysis of InSAR time-series. Structural Health Monitoring, 2022; 21(4), 1849-1878. https://doi.org/10.1177/14759217211045912

[8] Koohmishi M, Kaewunruen S, Chang L, Guo Y. Advancing railway track health monitoring: Integrating GPR, InSAR and machine learning for enhanced asset management. Automation in Construction, 2024, 162, 105378. https://doi.org/10.1016/j.autcon.2024.105378

[9] Giordano PF, Quqa S, Limongelli MP. The value of monitoring a structural health monitoring system. Structural Safety, 2023; 100, 102280. https://doi.org/10.1016/j.strusafe.2022.102280

[10] Catbas N, Avci O. A review of latest trends in bridge health monitoring. In Proceedings of the Institution of Civil Engineers-Bridge Engineering, 2022; 176(2), 76-91. https://doi.org/10.1680/jbren.21.00093

[11] Dong C, Li L, Yan J, Zhang Z, Pan H, Catbas F N. Pixel-level fatigue crack segmentation in large-scale images of steel structures using an encoder–decoder network. Sensors, 2021; 21(12), 4135. https://doi.org/10.3390/s21124135

[12] He Z, Chen W, Zhang J, Wang Y H. Crack segmentation on steel structures using boundary guidance model. Automation in Construction, 2024, 162, 105354. https://doi.org/10.1016/j.autcon.2024.105354

[13] Nettis A, Massimi V, Nutricato R, Nitti DO, Samarelli S, Uva G. Satellite-based interferometry for monitoring structural deformations of bridge portfolios. Automation in Construction, 2023; 147, 104707. https://doi.org/10.1016/j.autcon.2022.104707

[14] Liu W, Chen SE, Hauser E. Remote sensing for bridge health monitoring. In Atmospheric and Environmental Remote Sensing Data Processing and Utilization V: Readiness for GEOSS III, 2009; 7456, 100-109. SPIE. https://doi.org/10.1117/12.825528

[15] Selvakumaran S, Plank S, Geiß C, Rossi C, Middleton C. Remote monitoring to predict bridge scour failure using Interferometric Synthetic Aperture Radar (InSAR) stacking techniques. International journal of applied earth observation and geoinformation, 2018; 73, 463-470. https://doi.org/10.1016/j.jag.2018.07.004

[16] Karimzadeh S, Ghasemi M, Matsuoka M, Yagi K, Zulfikar AC. A deep learning model for road damage detection after an earthquake based on synthetic aperture radar (SAR) and field datasets. IEEE Journal of Selected Topics in Applied Earth Observations and Remote Sensing, 2022; 15, 5753-5765. https://doi.org/10.1109/JSTARS.2022.3189875

[17] Farneti E, Cavalagli N, Venanzi I, Salvatore W, Ubertini F. Residual service life prediction for bridges undergoing slow landslide-induced movements combining satellite radar interferometry and numerical collapse simulation. Engineering Structures, 2023; 293, 116628. https://doi.org/10.1016/j.engstruct.2023.116628

[18] Boloorani AD, Darvishi M, Weng Q, Liu X. Post-war urban damage mapping using InSAR: the case of Mosul City in Iraq. ISPRS International Journal of Geo-Information, 2021; 10(3), 140. https://doi.org/10.3390/ijgi10030140

[19] Wen D, Huang X, Bovolo F, Li J, Ke X, Zhang A, Benediktsson JA. Change detection from very-high-spatial-resolution optical remote sensing images: Methods, applications, and future directions. IEEE Geoscience and Remote Sensing Magazine, 2021; 9(4), 68-101. https://doi.org/10.1109/MGRS.2021.3063465

[20] Gbadamosi A Q, Oyedele L O, Delgado J M D, Kusimo H, Akanbi L, Olawale O, Muhammed-Yakubu N. IoT for predictive assets monitoring and maintenance: An implementation strategy for the UK rail industry. Automation in Construction, 2021; 122, 103486. https://doi.org/10.1016/j.autcon.2020.103486

[21] Schlögl M, Widhalm B, Avian M. Comprehensive time-series analysis of bridge deformation using differential satellite radar interferometry based on Sentinel-1. ISPRS Journal of Photogrammetry and Remote Sensing, 2021; 172, 132-146. https://doi.org/10.1016/j.isprsjprs.2020.12.001

[22] Blikharskyy Y, Kopiika N, Khmil R, Selejdak J, Blikharskyy Z. Review of development and application of digital image correlation method for study of stress–strain state of RC Structures. Applied Sciences, 2022; 12(19), 10157. https://doi.org/10.3390/app121910157

[23] Xiong S, Deng Z, Zhang B, Wang C, Qin X, Li Q. Deformation Evaluation of the South-to-North Water Diversion Project (SNWDP) Central Route over Handan in Hebei, China, Based on Sentinel-1A, Radarsat-2, and TerraSAR-X Datasets. Remote Sensing, 2023; 15(14), 3516. https://doi.org/10.3390/rs15143516

[24] Nikolakopoulos KG, Kyriou A, Koukouvelas IK, Tomaras N, Lyros E. UAV, GNSS, and InSAR Data Analyses for Landslide Monitoring in a Mountainous Village in Western Greece. Remote Sensing, 2023; 15(11), 2870. https://doi.org/10.3390/rs15112870

[25] Yoon H, Shin J, Spencer Jr B F. Structural displacement measurement using an unmanned aerial system. Computer-Aided Civil and Infrastructure Engineering, 2018; 33(3), 183-192. https://doi.org/10.1111/mice.12338

[26] Davila Delgado JM, Butler LJ, Gibbons N, Brilakis I, Elshafie MZ, Middleton C. Management of structural monitoring data of bridges using BIM. In Proceedings of the Institution of Civil Engineers-Bridge Engineering, 2017; 170(3), 204-218. https://doi.org/10.1680/jbren.16.00013

[27] Braik A M, Koliou M.Automated building damage assessment and large-scale mapping by integrating satellite imagery, GIS, and deep learning. Comput Aided Civ Inf., 2024;1–16. https://doi.org/10.1111/mice.13197

[28] Markogiannaki O, Chen F, Xu H, Mitoulis SA, Parcharidis I. Monitoring of a landmark bridge using SAR Interferometry coupled with engineering data and forensics. International Journal of Remote Sensing, 2022; 43(1) :95-119. https://doi.org/10.1080/01431161.2021.2003468

[29] Giordano PF, Turksezer ZI, Previtali M, Limongelli MP. Damage detection on a historic iron bridge using satellite DInSAR data. Structural Health Monitoring, 2022; 21(5), 2291-2311. https://doi.org/10.1177/14759217211054350

[30] Peduto D, Giangreco C, Venmans AA. Differential settlements affecting transition zones between bridges and road embankments on soft soils: Numerical analysis of maintenance scenarios by multi-source monitoring data assimilation. Transportation Geotechnics, 2020; 24, 100369. https://doi.org/10.1016/j.trgeo.2020.100369

[31] Alexander QG, Hoskere V, Narazaki Y, Maxwell A, Spencer Jr BF. Fusion of thermal and RGB images for automated deep learning based crack detection in civil infrastructure. AI in Civil Engineering, 2022; 1(1), 3. https://doi.org/10.1007/s43503-022-00002-y

[32] Kim H, Sim SH, Spencer BF. Automated concrete crack evaluation using stereo vision with two different focal lengths. Automation in Construction, 2022; 135, 104136. https://doi.org/10.1016/j.autcon.2022.104136

[33] Terrados-Cristos M, Ortega-Fernández F, Díaz-Piloneta M, Montequín VR, González JG. Potential Structural Damage Characterization through Remote Sensing Data: A Nondestructive Experimental Case Study. Advances in Civil Engineering, 2022; 6557898. https://doi.org/10.1155/2022/6557898

[34] Dong C Z, Catbas F N. A review of computer vision–based structural health monitoring at local and global levels. Structural Health Monitoring, 2021; 20(2), 692-743. https://doi.org/10.1177/1475921720935585

[35] Dong C, Bas S, Catbas F N. Applications of computer vision-based structural monitoring on long-span bridges in Turkey. Sensors, 2023; 23(19), 8161. https://doi.org/10.3390/s23198161

[36] Smail T, Abed M, Mebarki A, Lazecky M. Earthquake-induced landslide monitoring and survey by means of InSAR. Natural Hazards and Earth System Sciences, 2022; 22(5), 1609-1625. https://doi.org/10.5194/nhess-22-1609-2022

[37] Papadopoulos GA, Agalos A, Karavias A, Triantafyllou I, Parcharidis I, Lekkas E. Seismic and Geodetic Imaging (DInSAR) Investigation of the March 2021 Strong Earthquake Sequence in Thessaly, Central Greece. Geosciences. 2021; 11(8):311. https://doi.org/10.3390/geosciences11080311

[38] Mavroulis S, Triantafyllou I, Karavias A, Gogou M, Katsetsiadou K-N, Lekkas E, Papadopoulos GA, Parcharidis I. Primary and Secondary Environmental Effects Triggered by the 30 October 2020, Mw = 7.0, Samos (Eastern Aegean Sea, Greece) Earthquake Based on Post-Event Field Surveys and InSAR Analysis. Applied Sciences. 2021; 11(7):3281. https://doi.org/10.3390/app11073281

[39] Lekkas E, Mavroulis S, Gogou M, Papadopoulos GA, Triantafyllou I, Katsetsiadou KN, Kranis H, Skourtsos E, Carydis P, Voulgaris N, Papadimitriou P, Kapetanidis V, Karakonstantis A, Spingos I, Kouskouna V, Kassaras I, Kaviris G, Pavlou K, Sakkas V, Karatzetzou A, Evelpidou N, Karkani E, Kampolis I, Nomikou P, Lambridou D, Krassakis P, Foumelis M, Papazachos C, Karavias A, Bafi D, Gatsios T, Markogiannaki O, Parcharidis I, Ganas A, Tsironi V, Karasante I, Galanakis D, Kontodimos K, Sakellariou D, Theodoulidis N, Karakostas C, Lekidis V, Makra K, Margaris V, Morfidis K, Papaioannou C, Rovithis E, Salonikios T, Kourou A, Manousaki M, Thoma T, Karveleas N.(2020). The October 30, 2020 Mw 6.9 Samos (Greece) earthquake. Newsletter of Environmental, Disaster and Crises Management Strategies, 21, ISSN 2653-9454

[40] Baek W K, Jung H S. Precise three-dimensional deformation retrieval in large and complex deformation areas via integration of offset-based unwrapping and improved multiple-aperture SAR interferometry: application to the 2016 Kumamoto earthquake. Engineering, 2020; 6(8), 927-935. https://doi.org/10.1016/j.eng.2020.06.012

[41] Wang Y, Chew A W Z, Zhang L. Building damage detection from satellite images after natural disasters on extremely imbalanced datasets. Automation in construction, 2022, 140, 104328. https://doi.org/10.1016/j.autcon.2022.104328

[42] Tavakkoliestahbanati A, Milillo P, Kuai H, Giardina G. Pre-collapse spaceborne deformation monitoring of the Kakhovka dam, Ukraine, from 2017 to 2023. Communications Earth & Environment, 2024; 5(1), 145. https://doi.org/10.1038/s43247-024-01284-z

[43] Bacastow TS, Bellafiore DJ. Redefining Geospatial Intelligence. Am. Intell. J. 2009; 27, 38–40. https://www.jstor.org/stable/44327109

[44] Krassakis P, Karavias A, Nomikou P, Karantzalos K, Koukouzas N, Kazana S, Parcharidis I. Geospatial Intelligence and Machine Learning Technique for Urban Mapping in Coastal Regions of South Aegean Volcanic Arc Islands. Geomatics. 2022; 2(3):297-322. https://doi.org/10.3390/geomatics2030017

[45] Markogiannaki O, Karavias A, Bafi D, Angelou D, Parcharidis I. A geospatial intelligence application to support post-disaster inspections based on local exposure information and on co-seismic DInSAR results: the case of the Durres (Albania) earthquake on November 26, 2019. Nat Hazards, 2020; 103, 3085–3100. https://doi.org/10.1007/s11069-020-04120-7

[46] Krassakis P, Karavias A, Nomikou P, Karantzalos K, Koukouzas N, Kazana S, Parcharidis I. Geospatial Intelligence and Machine Learning Technique for Urban Mapping in Coastal Regions of South Aegean Volcanic Arc Islands. Geomatics. 2022; 2(3):297-322. https://doi.org/10.3390/geomatics2030017

[47] Krassakis P, Karavias A, Nomikou P, Karantzalos K, Koukouzas N, Athinelis I, Kazana S, Parcharidis I. Multi-Hazard Susceptibility Assessment Using the Analytical Hierarchy Process in Coastal Regions of South Aegean Volcanic Arc Islands. GeoHazards. 2023; 4(1):77-106. https://doi.org/10.3390/geohazards4010006

[48] Krassakis P, Kazana S, Chen F, Koukouzas N, Parcharidis I, Lekkas E. Detecting subsidence spatial risk distribution of ground deformation induced by urban hidden streams. Geocarto International, 2019; 1–13. https://doi.org/10.1080/10106049.2019.1622601

[49] Yun SH, Hudnut K, Owen S, Webb F, Simons M, Sacco P, Gurrola E, Manipon G, Liang C, Fielding E, Milillo P, Hua H, Colettaet A. Rapid damage mapping for the 2015 Mw 7.8 Gorkha earthquake using synthetic aperture radar data from COSMO-SkyMed and ALOS-2 satellites. Seismol Res Lett, 2015; 86(6):1549–1556. https://doi.org/10.1785/0220150152

[50] Sharma RC, Tateishi R, Hara K, Thanh Nguyen H, Gharechelou S, Viet Nguyen L. Earthquake Damage Visualization (EDV) technique for the rapid detection of earthquake-induced damages using SAR data. Sensors, 2017; 17(2) https://doi.org/10.3390/s17020235

[51] Preiss M, Gray DA, Stacy NJ. Detecting scene changes using synthetic aperture radar interferometry. IEEE Trans. Geosci. Remote Sens., 2006; 44(8), 2041-2054. https://doi.org/10.1109/TGRS.2006.872910.

[52] Bouaraba A, Younsi A, Belhadj Aissa A, Acheroy M, Milisavljevic N, Closson D. Robust techniques for coherent change detection using COSMO-SkyMed SAR images, Progr. Electromagn. Res., 2012, 22, 219-232, https://doi.org/10.2528/PIERM11110707 .

[53] ElGharbawi T, Zarzoura F. Damage detection using SAR coherence statistical analysis, application to Beirut, Lebanon. ISPRS Journal of Photogrammetry and Remote Sensing, 2021; 173, 1-9. https://doi.org/10.1016/j.isprsjprs.2021.01.001

[54] Yin J, Dong J, Hamm N A S, Li Z, J. Wang J, Xing H, Fu P. Integrating remote sensing and geospatial big data for urban land use mapping: A review. International Journal of Applied Earth Observation and Geoinformation, 2021; 103, 102514. https://doi.org/10.1016/j.jag.2021.102514

[55] Hermosilla T, Wulder M A, White J C, Coops N C. Land cover classification in an era of big and open data: Optimizing localized implementation and training data selection to improve mapping outcomes. Remote Sensing of Environment, 2022; 268, 112780. https://doi.org/10.1016/j.rse.2021.112780

[56] Gong P, W. Zhang W, Yu L, Li C, Wang J, Liang L, Li X, Ji L, Bai Y. New research paradigm for global land cover mapping. National Remote Sensing Bulletin, 2016; 20(5), 1002-1016. https://doi.org/10.11834/jrs.20166138

[57] Spencer Jr B F, Hoskere V, Narazaki Y. Advances in computer vision-based civil infrastructure inspection and monitoring. Engineering, 2019; 5(2), 199-222. https://doi.org/10.1016/j.eng.2018.11.030

[58] Guo J, Liu P, Xiao B, Deng L, Wang Q. Surface defect detection of civil structures using images: Review from data perspective. Automation in Construction, 2024,158, 105186. https://doi.org/10.1016/j.autcon.2023.105186

[59] Sacks R, Girolami M, Brilakis I. Building information modelling, artificial intelligence and construction tech. Developments in the Built Environment, 2020; 4, 100011. https://doi.org/10.1016/j.dibe.2020.100011

[60] Ren K, Zheng T, Qin Z, Liu X. Adversarial attacks and defenses in deep learning. Engineering, 2020; 6(3), 346-360. https://doi.org/10.1016/j.eng.2019.12.012

[61] Xing E P, Ho Q, Xie P, Wei D. Strategies and principles of distributed machine learning on big data. Engineering, 2016; 2(2), 179-195. https://doi.org/10.1016/J.ENG.2016.02.008

[62] Zhu Q, Lei Y, Sun, X, Guan Q, Zhong Y, Zhang L, Li D. Knowledge-guided land pattern depiction for urban land use mapping: A case study of Chinese cities. Remote Sensing of Environment, 2022; 272, 112916. https://doi.org/10.1016/j.rse.2022.112916

[63] Koch C, Brilakis I. Pothole detection in asphalt pavement images. Advanced engineering informatics, 2011; 25(3), 507-515. https://doi.org/10.1016/j.aei.2011.01.002

[64] Sertel E, Ekim B, Osgouei P E, Kabadayi M E. MaLand Use and Land Cover Mapping Using Deep Learning Based Segmentation Approaches and VHR Worldview-3 Images. Remote Sensing, 2022; 14(18), 4558. https://doi.org/10.3390/rs14184558

[65] Chong W. K., Naganathan H., Liu H., Ariaratnam S., Kim, J. Understanding infrastructure resiliency in Chennai, India using Twitter's Geotags and texts: a preliminary study. Engineering, 2018, 4(2), 218-223. https://doi.org/10.1016/j.eng.2018.03.010

[66] Narazaki Y, Hoskere V, Hoang T A., Fujino Y, Sakurai A, Spencer Jr B F. Vision-based automated bridge component recognition with high-level scene consistency. Computer-Aided Civil and Infrastructure Engineering, 2020; 35(5), 465-482. https://doi.org/10.1111/mice.12505

[67] Xiong C, Li Q, Lu X. Automated regional seismic damage assessment of buildings using an unmanned aerial vehicle and a convolutional neural network. Automation in Construction, 2020; 109, 102994. https://doi.org/10.1016/j.autcon.2019.102994

[68] Yang X, del Rey Castillo E, Zou Y, Wotherspoon L, Tan Y. Automated semantic segmentation of bridge components from large-scale point clouds using a weighted superpoint graph. Automation in Construction, 2022, 142, 104519. https://doi.org/10.1016/j.autcon.2022.104519

[69] LeCun Y, Bengio Y, Hinton G. Deep learning. Nature, 2015; 521(7553), 436-444. https://doi.org/10.1038/nature14539

[70] Deng J, Singh A, Zhou Y, Lu Y, Lee VCS. Review on computer vision-based crack detection and quantification methodologies for civil structures. Construction and Building Materials, 2022; 356, 129238. https://doi.org/10.1016/j.conbuildmat.2022.129238

[71] Pan Y, Braun A, Brilakis I, Borrmann A. Enriching geometric digital twins of buildings with small objects by fusing laser scanning and AI-based image recognition. Automation in Construction, 2022; 140, 104375. https://doi.org/10.1016/j.autcon.2022.104375

[72] Zakaria M, Karaaslan E, Catbas F N. Advanced bridge visual inspection using real-time machine learning in edge devices. Advances in Bridge Engineering, 2022; 3(1), 1-18. https://doi.org/10.1186/s43251-022-00073-y

[73] Chow J K, Liu K F, Tan P S, Su Z, Wu J, Li Z, Wang Y H. Automated defect inspection of concrete structures. Automation in Construction, 2021; 132, 103959. https://doi.org/10.1016/j.autcon.2021.103959

[74] Vaswani A, Shazeer N, Parmar N, Uszkoreit J, Jones L, Gomez A N, Kaise Ł, Polosukhin I. Attention is all you need. Advances in neural information processing systems, 2017; 30.

[75] Brown T, Mann B, Ryder, N, Subbiah M. Kaplan J. D, Dhariwal P, Pranav Shyam N, Sastry G, Askell A, Agarwal S, Herbert-Voss A, Krueger G, Henighan T, Child R, Ramesh A, Ziegler D, Wu J, Winter C, Hesse C, Chen C, Sigler E, Litwin M, Gray S, Chess B, Clark J, Berner C, McCandlish S, Radford A, Sutskever I, Amodei D. Language models are few-shot learners. Advances in neural information processing systems, 2020; 33, 1877-1901.

[76] Drobnyi V, Li S, Brilakis I. Deep-learning guided structural object detection in large-scale, occluded indoor point cloud datasets. 2023 European Conference on Computing in Construction. Heraklion, Crete, Greece. July 10-12, 2023.

[77] Du F, Jiao S, Chu K. Application research of bridge damage detection based on the improved lightweight convolutional neural network model. Appl. Sci. 2022; 12, 6225. https://doi.org/10.3390/app12126225

[78] Hou F, Lei W, Li S, Xi J, Xu M, Luo J. Improved Mask R-CNN with distance guided intersection over union for GPR signature detection and segmentation. Automation in Construction, 2021; 121, 103414. https://doi.org/10.1016/j.autcon.2020.103414

[79] Zhang Y, Yuen K-V. Review of artificial intelligence-based bridge damage detection. Advances in Mechanical Engineering, 2022; 14(9). https://doi.org/10.1177/16878132221122770 .

[80] Mission ends for Copernicus Sentinel-1B satellite. Available at: https://www.esa.int/Applications/Observing_the_Earth/Copernicus/Sentinel-1/Mission_ends_for_Copernicus_Sentinel-1B_satellite [accessed 3 March 2024].

[81] Torres R, Snoeij P, Geudtner D, Bibby D, Davidson M, Attema E, Potin P, Rommen BO, Floury N, Brown M, Traver IN, Deghaye P, Duesmann B, Rosich B, Miranda N, Bruno C, L'Abbate M, Croci R, Pietropaolo A, Huchler M, Rostan F, GMES Sentinel-1 mission. Remote Sens. Environ., 2012; 120, 9–24. https://doi.org/10.1016/j.rse.2011.05.028.

[82] Science Toolbox Exploitation Platform: https://step.esa.int/main/doc/tutorials/ [accessed 11 March 2024]

[83] OpenStreetMap. Available online: https://www.openstreetmap.org/#map=9/36.6904/25.0516 [accessed on 19 September 2023].

[84] Online source Damage In UA https://damaged.in.ua/about [accessed 3 March 2024]

[85] Online source: The Eyes on Russia. The Centre for Information Resilience (CIR). https://eyesonrussia.org/about ure%2CFood+and+agriculture&dateRange=1644789600000%2C1695675600000&onlyEventsMapFrame=false [accessed 3 October 2023]

[86] Online source: UADamage https://www.uadamage.com/map?h=MTIsMzguMDAxNjYsNDguNTkwNTU= [accessed 3 October 2023]

[87] USGS EROS Archive—Digital Elevation—Shuttle Radar Topography Mission (SRTM) 1 Arc-Second Global. U.S. Geological Survey. Available online: https://www.usgs.gov/centers/eros/science/usgs-eros-archive-digital-elevation-shuttle-radar-topography-mission-srtm-1 [accessed 3 March 2024].

[88] Closson D, Milisavljevic N. InSAR Coherence and Intensity Changes Detection. Mine Action-The Research Experience of the Royal Military Academy of Belgium, 2017; 292. https://doi.org/10.5772/65779

[89] Stramondo S, Bignami C, Chini M, Pierdicca N, Tertulliani A. Satellite radar and optical remote sensing for earthquake damage detection: results from different case studies. International Journal of Remote Sensing, 2006; 27(20), 4433-4447. https://doi.org/10.1080/01431160600675895

[90] Jung J, Yun S-H. Evaluation of Coherent and Incoherent Landslide Detection Methods Based on Synthetic Aperture Radar for Rapid Response: A Case Study for the 2018 Hokkaido Landslides. Remote Sensing. 2020; 12(2):265. https://doi.org/10.3390/rs12020265

[91] Kirillov A, Mintun E, Ravi N, Mao H, Rolland C, Gustafson L, Xiao T, Whitehead S, Berg AC, Lo WY, Dollár P, Girshick R Segment anything, 2023; arXiv preprint arXiv:2304.02643. https://doi.org/10.48550/arXiv.2304.02643

[92] Chen K, Liu C, Chen H, Zhang H, Li W, Zou Z, Shi Z. RSPrompter: Learning to prompt for remote sensing instance segmentation based on visual foundation model. arXiv:2306.16269v2, 2023; https://doi.org/10.48550/arXiv.2306.16269

[93] Liu S, Zeng Z, Ren T, Li F, Zhang H, Jie Y, Li C, Yang J, Su H, Zhu J, Zhang L. Grounding DINO: Marrying DINO with Grounded Pre-Training for Open-Set Object Detection. arXiv:2303.05499, 2023; https://doi.org/10.48550/arXiv.2303.05499

[94] Li L, Zhang P, Zhang H, Yang J, Li C, Zhong Y, Wang L, Yuan L, Zhang L, Hwang J, Chang KW, Gao, J. Grounded Language-Image Pre-training. Proceedings of the IEEE/CVF Conference on Computer Vision and Pattern Recognition (CVPR), 2022; 10965-10975. https://doi.org/10.48550/arXiv.2112.03857

[95] Bach Khoa Ho Chi Minh University. Crack detection using instance segmentation in yolov8 dataset (https://universe.roboflow.com/bach-khoa-ho-chi-minh-university-fyr43/crack-detection-using-instance-segmentation-in-yolov8) [accessed 15 December 2023]

[96] Rombach R, Blattmann A, Lorenz D, Esser P, Ommer B. High-resolution image synthesis with latent diffusion models. Proceedings of the IEEE/CVF conference on computer vision and pattern recognition, 2022; 10684-10695.

[97] Television news service. http://surl.li/lzmkl, [accessed 3 March 2024]

[98] Daniel Rice. Ukrainian Bridges are Playing a Vital Role in Both the Defense and the Offense in the Ukrainian-Russian War https://smallwarsjournal.com/jrnl/art/ukrainian-bridges-are-playing-vital-role-both-defense-and-offense-ukrainian-russian-war [accessed 3 March 2024]

[99] Maria Engqvist. 2022. A Railhead Too Far: The Strategic Role of Railroads during Russia's Invasion of Ukraine (https://www.foi.se/rest-api/report/FOI%20Memo%207954) [accessed 3 March 2024]

[100] Online source: https://twitter.com/detresfa_/status/1501413574497153024 [accessed 3 March 2024]

[101] Online source: https://www.slidstvo.info/warnews/my-u-2014-rotsi-z-donetska-pishly-z-toyi-zh-prychyny-yak-ludy-evakuyuyutsya-z-irpenya/ [accessed 3 March 2024]

[102] Li Z, Zou W, Ding X, Chen Y, Liu G. A quantitative measure for the quality of InSAR interferograms based on phase differences. Photogrammetric Engineering & Remote Sensing, 2004, 70(10), 1131-1137. https://doi.org/10.14358/PERS.70.10.1131

[103] Guarnieri A M, Prati C. SAR interferometry: A" quick and dirty" coherence estimator for data browsing. IEEE Transactions on Geoscience and Remote Sensing,1997, 35(3), 660-669. https://doi.org/10.1109/36.581984 .

[104] Zou W, Li Y, Li Z, Ding X. Improvement of the accuracy of InSAR image co-registration based on tie points–a review. Sensors, 2009, 9(02), 1259-1281. https://doi.org/10.3390/s90201259